\documentclass[11pt]{article}

\usepackage[]{ACL2023}

\usepackage{times}
\usepackage{latexsym}

\usepackage[T1]{fontenc}

\usepackage[utf8]{inputenc}
\usepackage{ xcolor, colortbl} 

\usepackage{microtype}

\usepackage{graphicx}
\usepackage{inconsolata}

\usepackage{siunitx} 
\usepackage{tikz}

\newcommand{\model}{Freshbench}
\usepackage{caption}
\usepackage{amsthm}
\theoremstyle{plain}

\theoremstyle{claim}

\theoremstyle{fact}

\theoremstyle{definition}
\newtheorem{definition}{Definition}[section]
\theoremstyle{definition}
\newtheorem{hypo}{Hypothesis Testing}
\theoremstyle{lemma}

\usepackage{wrapfig,lipsum,booktabs}

\theoremstyle{proper}
\theoremstyle{definition}

\usepackage{adjustbox}   
\usepackage{rotating}    

\usepackage{subcaption}
\usepackage{amsmath}
\usepackage{amssymb}
\usepackage{bm} 
\usepackage{makecell}
\usepackage{siunitx}
\usepackage{geometry} 
\usepackage{array} 
\usepackage{multirow} 
\usepackage{ xcolor, colortbl} 

\usepackage[utf8]{inputenc} 
\usepackage[T1]{fontenc}    
\usepackage{hyperref}       
\usepackage{url}            
\usepackage{booktabs}       
\usepackage{amsfonts}       
\usepackage{nicefrac}       
\usepackage{microtype}      
\usepackage{xcolor}         

\usepackage[most]{tcolorbox}
\usepackage{float}
\usepackage{xspace}
\tcbset{
aibox/.style={
width=430pt,
top=10pt,
colback=white,
colframe=black,
colbacktitle=black,
enhanced,
center,
attach boxed title to top left={yshift=-0.12in,xshift=0.15in},
boxed title style={boxrule=0pt,colframe=white},
}
}
\newtcolorbox{AIbox}[2][]{aibox,title=#2,#1}

\definecolor{darkgreen}{rgb}{0.0, 0.5, 0.0}
\definecolor{maroon}{rgb}{0.5, 0.0, 0.0}
\definecolor{navy}{rgb}{0.0, 0.0, 0.5}
\definecolor{teal}{rgb}{0.0, 0.5, 0.5}

\title{\textit{Is Your LLM Outdated?} A Deep Look at Temporal Generalization}

\author{%
Chenghao Zhu$^{1,2}$, 
Nuo Chen$^{1}$, 
Yufei Gao$^{1}$,\\
Yunyi Zhang$^{1}$,
Prayag Tiwari$^{3}$,
\textbf{Benyou Wang$^{1,2}$}\thanks{%
Benyou is the corresponding author.}\\
$^1$ The Chinese University of Hong Kong, Shenzhen\\
$^2$ Shenzhen Research Institute of Big Data\\
$^3$ Halmstad University\\
\texttt{wangbenyou@cuhk.edu.cn} \\
}

\begin{document}
\maketitle

\begin{abstract}

The rapid advancement of Large Language Models (LLMs) has led to the development of benchmarks that consider temporal dynamics, however, there remains a gap in understanding how well these models can generalize across temporal contexts due to the inherent dynamic nature of language and information.
This paper introduces the concept of temporal generalization in LLMs, including bias in past and future generalizations. Then we introduce FreshBench, a new evaluation framework that employs fresh text and event prediction for assessing LLMs' temporal adaptability, ensuring the evaluation process free from data leakage and subjective bias. The experiment shows significant temporal biases and a decline in performance over time.
Our findings reveal that powerful models, while initially superior, tend to decline more rapidly in future generalization. Additionally, powerful open-source models demonstrate better long-term adaptability compared to their closed-source counterparts. Our code is available at \url{https://github.com/FreedomIntelligence/FreshBench}

\end{abstract}


\section{Introduction}
Proprietary large language models (LLMs) 
~\citet{openai2023gpt4,bard,calude2,anil2023palm}
has been paralleled by efforts in the open-source community to democratize LLMs. Meanwhile, the rapid advancement of LLM democratization~\citet{touvron2023llama} emphasizes the necessity for dynamic and robust benchmarks that accurately reflect their evolving capabilities. Traditional benchmarks, broadly classified into knowledge-based assessments and open-dialogue evaluations, each present unique challenges. Knowledge-based assessments, exemplified by MMLU~\citet{hendrycks2020measuring} and C-Eval~\citet{huang2023ceval}, are prone to data manipulation, raising concerns about their real-world applicability and the risk of data leakage~\citet{wei2023skywork,golchin2024timetravelllmstracing}. 
Open-dialogue evaluations like MT-Bench~\citet{zheng2023judging} and Alpaca-Eval~\citet{alpaca_eval}, rely on subjective human or model-based judgments, making their outcomes susceptible to the influence of dialogue structure, and thereby, potentially compromising assessment validity.

Given the limitations of traditional methods, researchers are increasingly developing new approaches to evaluate models in diverse and dynamic contexts. For instance, recent efforts have introduced methods that assess models' abilities to handle temporal evolution, contextual understanding, and real-time information updates~\citet{fatemi2024tot, kasai2024realtimeqa, vu2023freshqa, chen2021timeqa, zhang21situatedqa, liška2022streamingqa}. While methods like RealTimeQA and FreshQA focus on continuously updating knowledge for real-time processing~\citet{kasai2024realtimeqa, vu2023freshqa}, others like LatestEval~\citet{LatestEval} emphasize recent text-based comprehension. In contrast, our work shifts the focus to evaluating models' world understanding through prediction, using a fact-based approach that minimizes bias and ensures objectivity in assessment. This method enhances robustness against data contamination and focuses on the model's capacity for accurate, real-world event prediction.

We believe that the evaluation of LLMs should be conducted in scenarios that are resistant to hacking manipulation, with results that are objective and \textbf{free from evaluator biases}~\citet{chen2024humans}. Additionally, time as a dynamic dimension, can also ensure robust and relevant evaluations. This paper explores an alternative way to \textbf{evaluate large language models}, i.e., \textbf{on temporal generalization} where time is considered as an important aspect.
This approach significantly reduces the issue of data contamination by limiting access to future content.
Moreover, a retrospective perspective on time could be beneficial for temporal bias analysis for LLMs.  

To achieve this goal, we design two scenarios. The first is to have the model generate text based on recent data such as arXiv papers, news articles, new books, and Wikipedia content. An excellent model should be able to generalize these new texts with high language likelihood without data leakage. The second scenario focuses on future event prediction. An excellent model should be able to understand the current context, integrate world knowledge, and effectively generalize to future scenarios. In these two scenarios, temporal generalization evaluation has two main advantages: eliminating the possibility of data leakage through temporal boundaries and enabling objective evaluation results without relying on subjective interpretations.

Our analysis indicates that using validation loss of new text to evaluate models is mostly effective. However, we observe that while performance on language likelihood is a strong indicator of a model's linguistic capabilities, it does not always align with specific abilities measured by existing benchmarks, which may represent discrete knowledge or other capacities that are important but not directly related to language modeling.

Our \textbf{c}\textbf{ontributions} are as follows:
1) We define and quantify temporal generalization and bias, establishing a basis for understanding and assessing the capabilities of LLMs over time.
2) We proposed a benchmark \model \ to test the capabilities of LLMs in predicting future events. This benchmark is dynamically updated to reflect the latest data, ensuring that our evaluations remain relevant and accurate in the face of rapidly evolving data environments.
It  experimentally assesses the performance of existing LLMs across various timeframes, providing insights into their historical and ongoing performance stability and adaptability.

\section{Conceptualization of Temporal Generalization}

In this section, we will first conceptualize \textit{Temporal Generalization}, then  \textit{Temporal Biases} and \textit{Temporal Degeneration} in Sec. \ref{sec:temporal_generalization}. We will discuss ways to measure /textit{Temporal Biases and Degeneration} in \ref{sec:assessing}.

\subsection{Temporal Generalization}
\label{sec:temporal_generalization}

LLMs face the ongoing challenge of remaining applicable as new text evolves. Consequently, it is crucial for these models to not only understand and generate text based on  seen (e.g., past) data but also to anticipate and adapt to future data. This inspires  temporal generalization as below.

\begin{figure}[h!]
    \centering
    \includegraphics[width=0.48\textwidth]{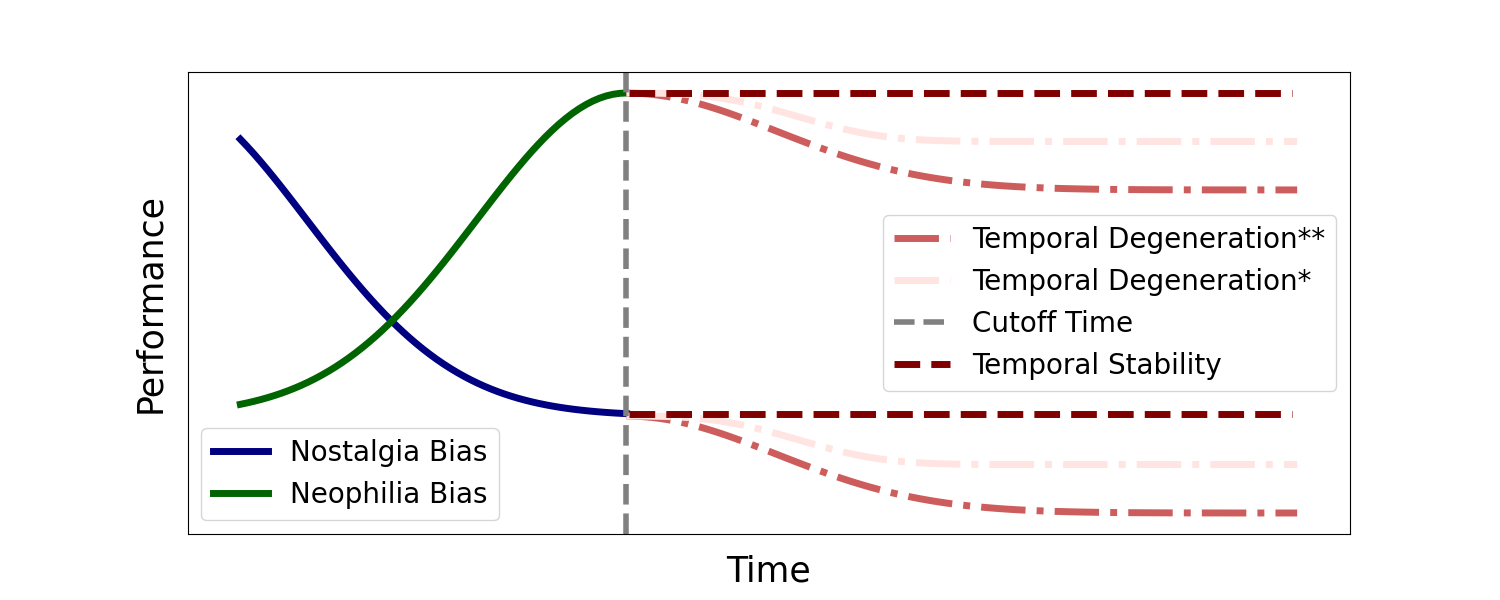}
\caption{Schematic figure of temporal generalization. In general,  prior to the cutoff time, an LLM might exhibit a \textcolor{navy}{Nostalgia} bias that is skewed toward past dates (i.e., aligning better with historical data) or a \textcolor{darkgreen}{Neophilia} bias that is skewed toward future dates ( favoring more recent information). However, after the cutoff time, the model's understanding of future trends declines, leading to temporal degeneration. }
\label{fig:vistal_bias}
\vspace{-4mm}
\end{figure}
\vspace{-2mm}

\vspace{1mm}
\begin{definition}
\textbf{Temporal Generalization} refers to the LLMs' ability to align with the contexts of the past, present, and future.
\end{definition}

\vspace{-0.8mm}
This requires the models to integrate and apply knowledge from historical and current data to adapt to the evolution of language and emerging trends. Temporal generalization highlights the challenge of designing models that remain effective and accurate over time, without the need for frequent re-training as new data is introduced. LLMs must maintain a delicate balance between leveraging historical data to understand long-standing language patterns and adapting to new expressions, terminology, and topics as they emerge.

\subsubsection{For Models: Define Past, Present and Future }
\begin{figure}[h!]
    \centering
    \vspace{-2mm}
    \includegraphics[width=0.48\textwidth]{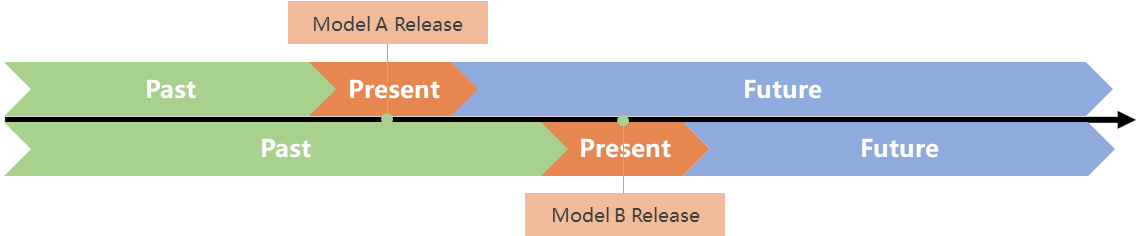}
    \vspace{-5mm}
    \caption{Schematic Diagram of 3 Periods of Models}
    \label{fig:timeline}
    \vspace{-3mm}
\end{figure}

To distinguish the \textit{past} and the \textit{future}, we define the cutoff time as the release date of the model. For instance, if an LLM is released in January 2024, then all events or information before January 2024 are considered part of the \textit{past}, while everything after January 2024 is considered the \textit{future}. Although the data after the cutoff time is not strictly future data, this approach ensures no data leakage for the model. As is shown in Figure \ref{fig:timeline}, the period near the cutoff time can be referred to as the \textit{present}. We evaluate the model's prediction accuracy across these time periods, denoted as $acc_{\text{past}}$, $acc_{\text{future}}$, and $acc_{\text{present}}$, respectively.

\label{3_period_def}

\subsubsection{Define Temporal Bias and Temporal Degeneration }

With the three time periods defined, we can define Temporal Bias and Temporal Degeneration for models.

\begin{definition}

\textbf{Temporal Bias} In this paper, we define temporal bias as the preferences of LLMs over time. \footnote{Temporal bias are multifaceted in previous literature. For instance, can refer to the assumption of an incorrect sequence of events that misleads our reasoning about causality~\citet{dorsey21possible} or user preferences over time in point-of-interest recommendations~\citet{Rahmani2022TemporalBias}. Recently, biases have raised more challenges in pre-trained language models~\citet{yogarajan2023tackling}, particularly focusing on under-represented societies. \citet{sarkar2024lookahead} examines lookahead bias in pre-trained language models, providing insights into how these biases affect model performance.}
\end{definition}

\label{sec:biases}

There exist two biases, namely \textbf{\textcolor{navy}{Nostalgia} Bias} and \textbf{\textcolor{darkgreen}{Neophilia} Bias}, which reflect two contrasting tendencies in how these models might predict or generate text about the past, present, or future. 
A visual representation of these biases is provided in Fig.~\ref{fig:vistal_bias}, where the trends associated with \textcolor{navy}{Nostalgia} and \textcolor{darkgreen}{Neophilia} biases are plotted over time.

    \textbf{\textcolor{navy}{Nostalgia} Bias} 
     refers to the over-reliance on historical data when generating text. This bias causes models to reflect perspectives, styles, or information prevalent in past training data, leading to outputs that may not account for recent developments. Essentially, \textcolor{navy}{Nostalgia} Bias can make a model seem `stuck in the past', limiting its ability to accurately represent the present or predict the future.

    \label{def:Nostalgia}
    \textbf{\textcolor{darkgreen}{Neophilia} Bias} implies a model's overemphasis on novelty and future trends, potentially at the expense of established information. This bias could result in predictions that are overly speculative and not well-grounded in historical data or current realities. Though less common due to LLM training on historical datasets, \textcolor{darkgreen}{Neophilia} Bias could be perceived from prompt structuring or interpretation of outputs that prioritize futuristic concepts over historical accuracy.

    \label{def:Neophilia}

\begin{definition}
\textbf{Temporal Degeneration} refers to the decline in a model's performance in the future,  that is, after it is released. 
\end{definition}

In conclusion, the illustration for the above concepts is in Figure \ref{fig:vistal_bias}.
\label{sec:define_bias&degeneration}

\subsection{Assessing Temporal Generalization}
\label{sec:assessing}

Based on definition in  Sec.~\ref{sec:temporal_generalization},  we  turn to assess temporal generalization  through the lens of 1) I:  compression intelligence in Sec.~\ref{sec:case1_likelihood} and II): future event prediction in Sec.~\ref{sec:prognostication}.

\subsubsection{Case I:  Compression Intelligence}
\label{sec:case1_likelihood}

In the assessment of compression intelligence, metrics such as perplexity (PPL) and bits per character (BPC) offer insights into model performance. 
We refine our focus to BPC for a more normalized comparison. Unlike traditional BPC which is specific to character-level language models, our \textbf{adapted Bits Per Character (BPC)} metric offers a comparison between models by normalizing the log likelihood based on character count, irrespective of the tokenization method, allowing for a more \textbf{equitable assessment of model performance} across different languages and tokenization schemes, the BPC is calculated as follows:



\vspace{-3mm}
\begin{equation} 
\!BPC(T) \!=\! \frac{-\sum_{i=1}^{N} \log p(w_i | w_1, \ldots, w_{i-1})}{\text{len-utf-8}(T)}
\end{equation}

Here, $T$ represents the text being analyzed, $N$ is the number of tokens in $T$ (which may vary depending on the tokenizer), and $\text{len-utf-8}(T)$ is the length of $T$ when encoded in UTF-8, measured in characters. 
The tokens \( w_i \) correspond to segments of text which may vary depending on the tokenizer used. This variance is due to different tokenization methods splitting the text into tokens at different granularities, ranging from subword units to characters. In this paper, we investigate the compression intelligence proposed in \citet{huang24compression} in a temporal manner,

\label{sec:future}

\subsubsection{Case II: Future Event Prediction}
\label{sec:prognostication}
LLMs could perform \textit{future event prediction}, as seen in \citet{bonde2022gpt3metaculus} which employed GPT-3 in a few-shot setting to tackle binary questions from the Metaculus platform. Further, \citet{zou2022forecasting} enriched the discourse by introducing a comprehensive dataset that includes forecasting questions alongside a news corpus. This approach enhances model training by incorporating varied and relevant data contexts. Moreover, \citet{halawi2024approaching} showcased how integrating news content can significantly improve the accuracy of predictions in real-time scenarios.

Building on this foundation, our research seeks to explore the temporal generalization capabilities of LLMs:\textit{ how well they adapt and remain accurate as new data emerges and contexts evolve?}
As an example of using LLMs to predict political outcomes, we gather new factual data about future events, such as \texttt{"What will President Biden's approval rating be as of 7 June 2024, according to FiveThirtyEight?"}. To ensure objectivity in our predictions, answers are typically formatted as ranges within a prediction market scenario. For instance, President Biden's approval rating might be categorized into intervals such as \texttt{lower than 34\%}, \texttt{at least 34.0\%, but less than 36.0\%}, \texttt{at least 36.0\%, but less than 38.0\%}, up to \texttt{46.0\% or higher}.

However, since collecting actual future data is impossible, we adopt a retrospective study approach. We use data available after the event's question was posed and before the prediction market's closure, treating this as "future" data for model evaluation. This ongoing, online evaluation objectively assesses the model's predictive capability. We define prediction accuracy, denoted by \(Acc\), as follows: 
\vspace{-5mm}
\begin{equation}
    Acc = \frac{N_{\text{correct}}} {N_{\text{total}}}
\end{equation}
\vspace{-5mm}

where \(N_{\text{correct}}\) is the number of correct predictions among \(N_{\text{total}}\) future prediction questions. 
See. Sec.~\ref{sec:Temporal_Biases} for the collected future prediction question set over time.
This metric evaluates the effectiveness of LLMs in predicting future events.
This approach also allows continuous evaluation with the latest data, ensuring the assessment reflects the model's current capabilities and the evolving information landscape.

\section{Temporal Generalization  through the Lens of Compression Intelligence}
\label{sec:likelihood}

This section discusses how various models handle data from different periods, showcasing their performance stability or volatility.

\begin{figure*}[ht]
    \centering
    \begin{minipage}{0.32\textwidth}
        \includegraphics[width=\textwidth]{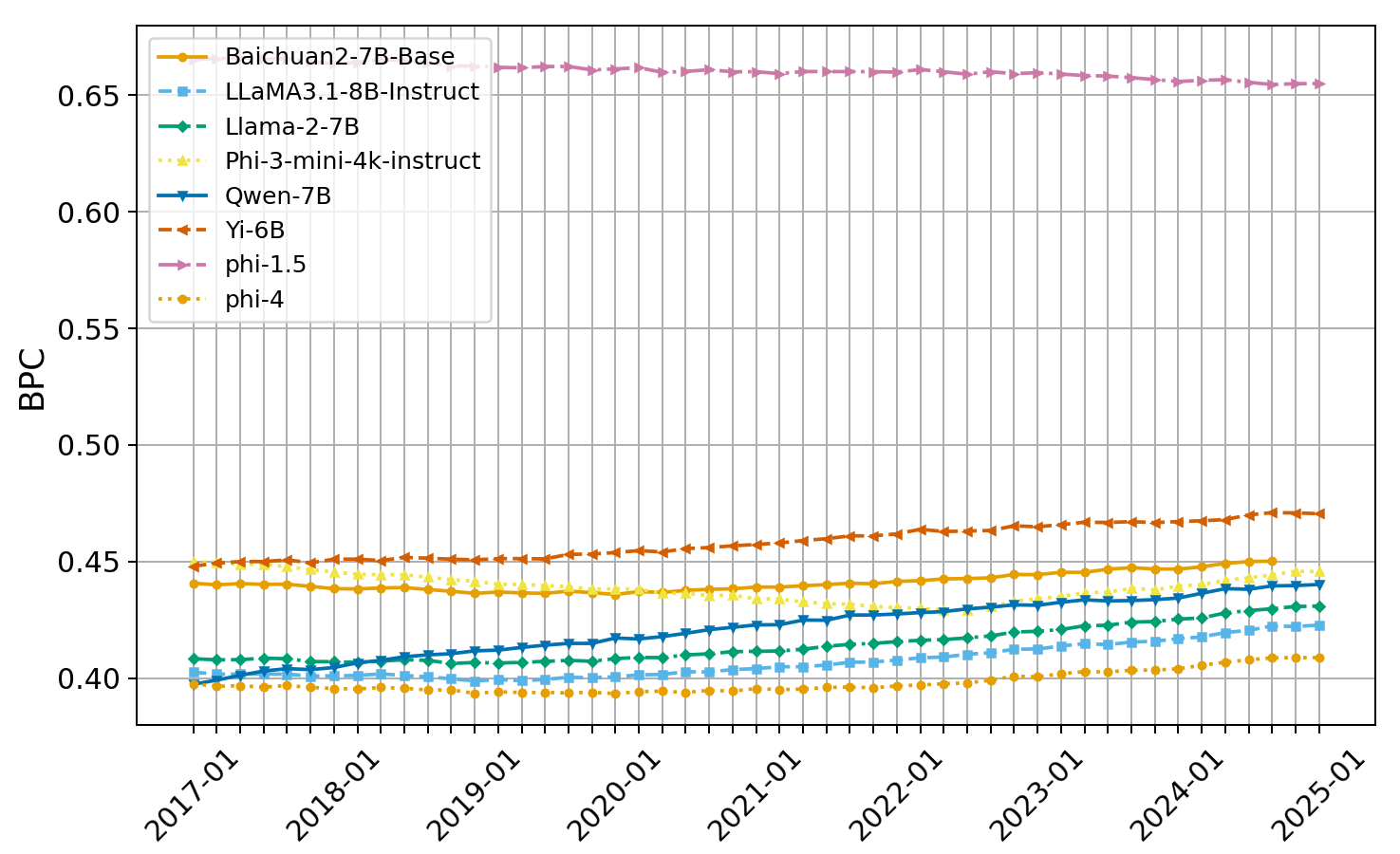}
        \caption{BPC on Wiki text}
        \label{fig:pplwiki}
    \end{minipage}\hfill
    \begin{minipage}{0.32\textwidth}
        \includegraphics[width=\textwidth]{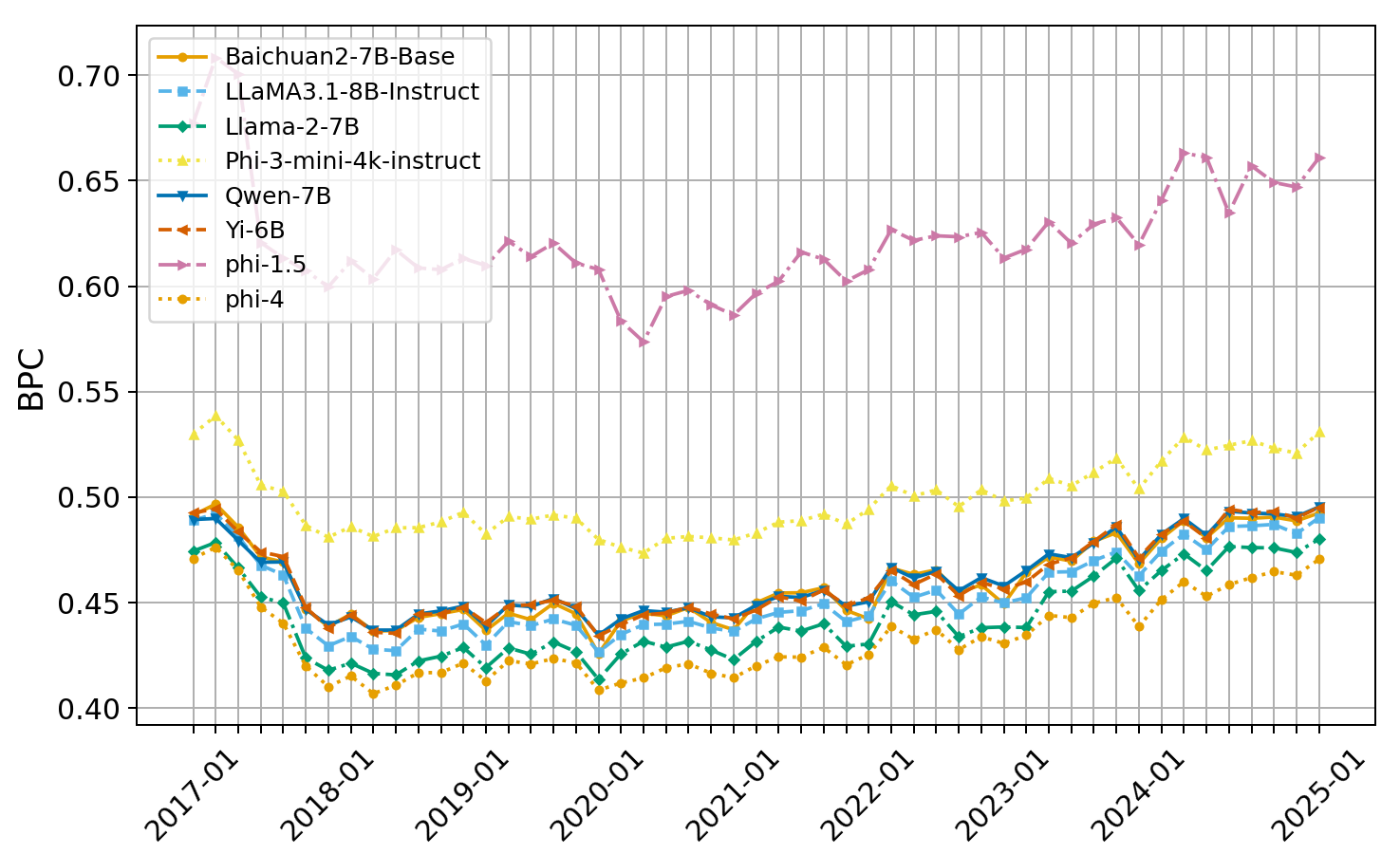}
        \caption{BPC on BBC news text}
        \label{fig:pplbbc}
    \end{minipage}
    \begin{minipage}{0.32\textwidth}
        \includegraphics[width=\textwidth]{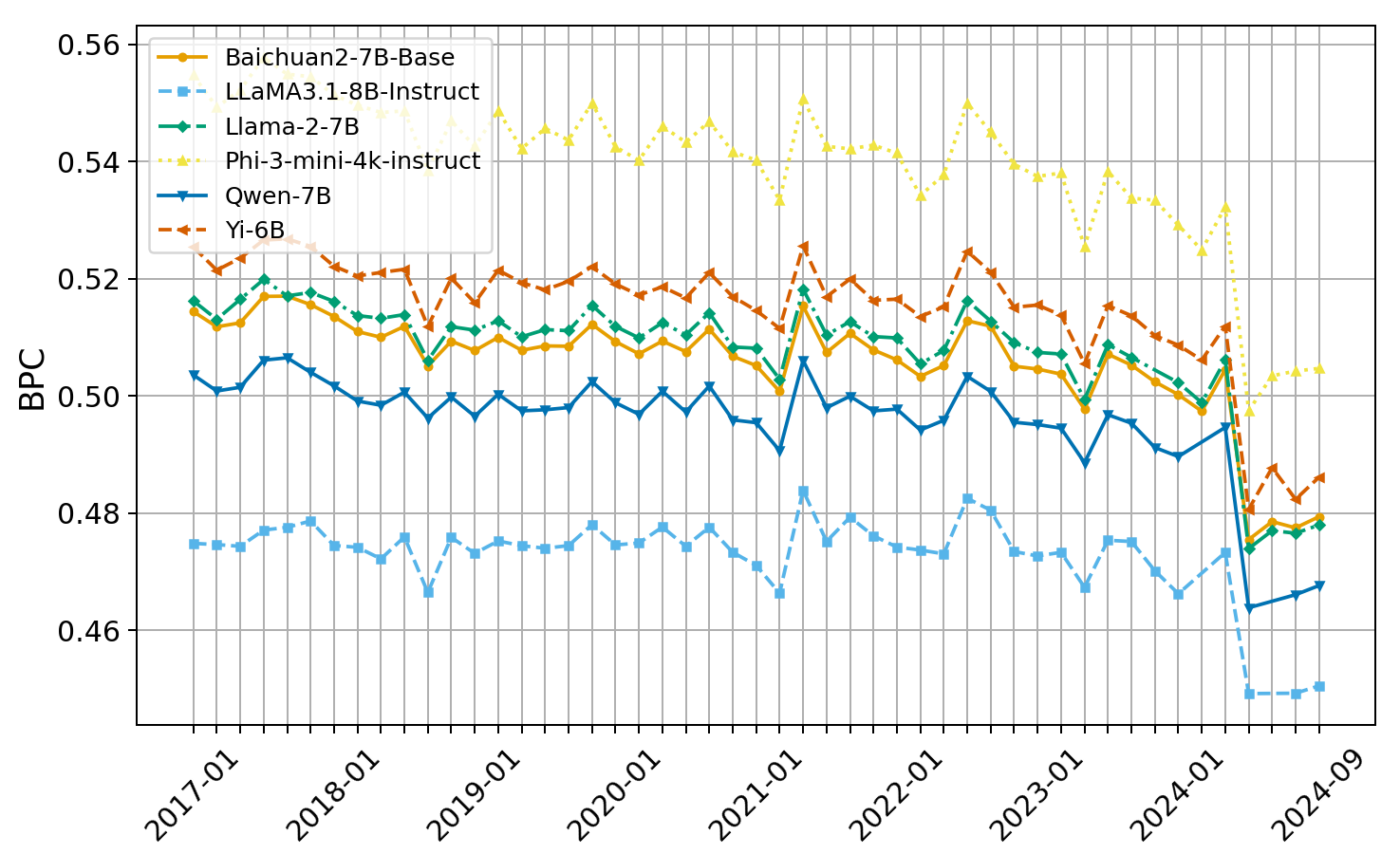}
        \caption{BPC on arXiv text}
        \label{fig:pplarxiv}
    \end{minipage}\hfill
    \vspace{-2mm}
\end{figure*}

\paragraph{Key Findings}
\vspace{-3mm}
\begin{itemize}
\item In general, while the ranking of models varied slightly between datasets, the performance of individual models remained relatively stable over time within each dataset type, as can be seen in Figure \ref{fig:pplwiki} and \ref{fig:pplbbc}. Here we made a discovery regarding the distinct characteristics of the Wikipedia and BBC datasets at \ref{Difference_of_Text_Feature}.

\item Another interesting observation was made in the arXiv dataset shown in Figure ~\ref{fig:pplarxiv}, where a sharp decrease in BPC occurred around March 2024. This timing coincides with the peak in ChatGPT usage, as reported by Exploding Topics\footnote{\url{https://explodingtopics.com/blog/chatgpt-users}}, which saw over 1.8 billion visits in that month.
In essence, it appears that humans may be adapting their writing to more closely resemble LLM outputs, rather than LLMs significantly enhancing their ability to model human language.
\label{arxiv_decrease}

\end{itemize}

\vspace{-3mm}
\subsection{Temporal Bias}

\subsubsection{Protocol for Quantitative Measurement}


\textbf{Data sources}
Our system aggregates data from various platforms, including Yahoo for financial news, BBC for political insights, Reddit for discussions, Wikipedia for encyclopedic updates, arXiv for academic research, GitHub for software trends, and Quora for diverse questions and answers. Utilizing a Python-based crawling framework, it's designed for adaptability and efficient data collection, with Playwright addressing navigation and specific site challenges. We also did some basic data pre-processing, due to space limitation, details in Appendix \ref{app:collection}. An overview of the data volume and the average entry length from a single crawl is provided in Appendix \ref{app:data_overview}.
Experiment settings are listed in Appendix \ref{app:exp2}.

\subsubsection{Observation of Temporal Bias in Compression Intelligence}

To comprehensively explore the temporal generalization capabilities of models, we plot the BPC curve over time. 

\textbf{In text data before release}, a negative slope indicates the presence of \textcolor{darkgreen}{Neophilia} Bias. Conversely, a positive slope would suggest that the model's performance is decreasing, which shows \textcolor{navy}{Nostalgia} Bias. 

ArXiv BPC has a huge decline after GPT is widely used, as mentioned in Key Findings. Models show stronger \textcolor{navy}{Nostalgia} Bias when evaluated with BBC text than Wiki text.
From the BPC charts, it's not easy to spot differences between models. However, our calculations in Appendix \ref{BPC_on_three_text_datasets} reveal variations in how well each model can adapt specific text over time. 

\textbf{Difference of Text Feature}
An insightful reviewer helped us found the reason of the different patterns of Wiki an BBC text. New Wiki texts with recent temporal markers do not necessarily imply that the content is new in a temporal context. For instance, in 2017-01, there was a Wiki article titled "History of India". However, the information within it was not new and could potentially have been sourced from other materials. Conversely, BBC texts are news articles and are highly time-sensitive.

To quantify the difference, we uniformly sampled approximately 500 data pieces from both the Wiki and BBC datasets (taking into account that news may also cover historical events) and asked gpt-4o to classify the content. The results are as follows:\\
\begin{itemize}
 \vspace{-6mm}
\item Wiki: Historical: 97.83\%, Others \footnote{"Others" refers to content not associated with specific historical events, focusing instead on general, educational, and ever-relevant information.} :2.17\%.
 \vspace{-2mm}
\item BBC: Recent: 91.86\%, Historical: 8.14\%.
\end{itemize}
 \vspace{-2mm}

So, it is reasonable that Wiki text has a flat curve in BPC.

\textbf{Claim the Value of Wiki Dataset} Since the Wikipedia dataset is continuously updated, it is not easily subject to direct contamination. Although a substantial portion of Wiki pertains to old information, this content still represents valuable knowledge that we need. An ideal model should consistently perform well on such knowledge. (If a model shows a decline in performance on Wikipedia, it is likely that the model is overfitting.)
\label{Difference_of_Text_Feature}

\subsection{Temporal Degeneration}

\subsubsection{Protocol for Quantitative Measurement}

\begin{definition}
    \textbf{Temporal Bias Index (TBI)} For each model in a certain period, we record $BPC_i$ at $n$ time points, forming $n$ data points $(i, BPC_i)$, where $i = 1, \ldots, n$. Here the time interval between $BPC_i$ and $BPC_{i+1}$ is set as two month. We then fit a linear model:

\begin{equation}
        BPC_i = a * i + b
\end{equation}

    then Temporal Bias Index (TBI) is defined as the coefficient $a$. TBI is crucial for understanding \textbf{the direction and rate of change in model performance over time}.
\end{definition}

To quantify model temporal generalization, we computed the BPC for various models on the Wiki and BBC datasets and evaluated their performance using this index. Specifically,
We established a base BPC using data from the six months prior to each model's release. After that, we measured the changes in BPC at 3, 6, 9 and 12 months post-release relative to the base BPC; and assessed the overall trend in model performance using TBI.
The specific results are detailed in Appendix \ref{appendix:ppl_generalization_wiki&BBC}. 

\begin{table*}[ht]
\small  
\caption{Correlation between Model Attributes and Their TBI on Wiki and BBC text}
\label{corr_bbc_wiki_bench_size_releaseDate}
\centering
\begin{tabular}{@{}p{0.9cm}@{} p{0.9cm}@{} p{1.1cm}@{} p{1.1cm}@{} p{1.1cm}@{} p{1.2cm}@{} p{1.1cm}@{} p{1.1cm}@{} p{0.9cm}@{} p{0.9cm}@{} p{0.9cm}@{} p{0.9cm}@{} p{1.0cm}@{} p{0.9cm}@{}}
\toprule
Corre & Model & Release & MMLU & Human- & Social- & STEM & MMLU & GSM8 & Ar20c & Hella- & Trust- & Wino- & Long- \\
lation & Size & Date &  & ities &  Sciences & & Other &  &  & wasg & fulQA & grande & bench \\
\midrule
Wiki & 0.504 & 0.267 & 0.599 & 0.500 & 0.594 & 0.393 & 0.528 & 0.473 & 0.380 & 0.581 & 0.200 & 0.632 & 0.404 \\
BBC & -0.044 & 0.313 & 0.384 & 0.317 & 0.311 & 0.186 & 0.153 & 0.188 & -0.071 & 0.248 & 0.485 & 0.204 & 0.549 \\
\bottomrule
\end{tabular}
    \vspace{-1mm}
\end{table*}

\subsubsection{Observation}

\textbf{Temporal generation pattern is mostly determined by the nature of data} Models perform stably on Wikipedia but show more variation on BBC. This reveals their ability to handle well-structured and formal text like Wiki and insufficient adaptability to dynamic text like BBC text. The overall BPC for BBC texts shows an increasing trend, indicating gradual performance degeneration.

Individual models show considerable differences in performance on new data; for instance, Yi, and Baichuan2 series adapt relatively well to new texts, while Phi-1.5 and OPT series exhibit poorer generalization (see Appendix \ref{BPC_on_three_text_datasets}). These findings indicate that in dynamic environments, different models have significantly varying adaptation capabilities.

\textbf{Model performance positively correlates with performance decay. (see Tab.~\ref{corr_bbc_wiki_bench_size_releaseDate}).} There is a positive correlation between model attributes (such as model size and release date) and performance decay. A positive BPC slope indicates an increase in BPC over time, reflecting performance degeneration.

In the Wiki dataset, larger models deteriorate faster (correlation coefficient of 0.504), suggesting a trade-off between capability and flexibility. In more dynamic environments (such as the BBC dataset), the impact of model size on performance decay is smaller. Additionally, newer models show higher correlations, meaning that despite their advanced capabilities, they may deteriorate faster. These results emphasize the importance of designing models that can maintain robustness over time and across various data environments. As machine learning applications grow in complexity and scale, temporal generalization will become increasingly critical.

\section{Temporal Generalization through the Lens of Future Event Prediction}
\label{sec:future}

\textbf{Advantages of Future Event Prediction over Compression Intelligence}
Compression Intelligence is limited by its inapplicability to closed-source models, where key indicators like logits or losses are inaccessible. This exclusion narrows comparative analysis and hinders a comprehensive understanding of the language technology landscape. Additionally, language likelihood can be biased by models finely tuned to specific formats or domains, skewing overall evaluation and presenting a distorted view of a model's general performance. 

This motivated us to research temporal generalization and bias from the perspective of future prediction.

In this section, we investigate temporal generalization (in Sec.~\ref{sec:Temporal_Biases}) and biases (Sec.~\ref{sec:Generalization}) through the Lens of Future Event Prediction.

\paragraph{Key Findings}
\begin{itemize}
    \item Models prefer older knowledge, with non-uniform knowledge distribution skewed towards the past. This poses a risk as users expect up-to-date knowledge close to the model's release date.
    \item Within model series such as Claude,Gemini, and GPT, newer versions are becoming more powerful but also experience more rapid performance decline. Miniature models significantly underperform compared to their full -sized counterparts in temporal tasks.
    \item Newer models generally exhibit stronger performance. Closed-source models tend to perform slightly better at the beginning, yet open-source models demonstrate superior generalization capabilities.
\end{itemize}


\subsection{Temporal Biases}
\label{sec:Temporal_Biases}

\subsubsection{Protocol for Quantitative Measurement: a Hypothesis Testing  Method}

\label{hyp_testing}

We discussed generally what is past, present, and future for a model in Sec \ref{3_period_def}. Now, let's clarify the context of the prediction problem. For each prediction task, there is an initiation time (when the question is posed) and a close time (when the answer is revealed). If a question's closure time precedes the model's release date (for instance, by more than three months), then that question is considered part of the \textit{past} for that model. The \textit{present} refers to questions whose close time is near the release time of the model, capturing the period around when the model was launched.

To investigate the two mentioned temporal biases in Sec.~\ref{sec:biases}, we propose two hypothesis testings as follows. These tests help practitioners make informed decisions despite the randomness inherent in the question datasets.
\begin{hypo}
\textbf{\textcolor{darkgreen}{Neophilia} Bias:} 
\begin{equation} 
    \begin{aligned}
        T_1: \quad & H_0: \mathbf{E}[acc_{\text{past}}] = \mathbf{E}[acc_{\text{present}}],\quad \\ 
                        & H_1: \mathbf{E}[acc_{\text{past}}] < \mathbf{E}[acc_{\text{present}}]\\
    \end{aligned}
\end{equation}

This hypothesis tests if there is a significant increase in model accuracy for current questions compared to past questions, suggesting a bias towards newer content. The p-value \( \widehat{p} \) quantifies the strength of evidence against the null hypothesis.
\end{hypo}

\begin{hypo}
\textbf{\textcolor{navy}{Nostalgia} Bias:} 
\begin{equation} 
    \begin{aligned}
        T_2: \quad & H_0^\prime: \mathbf{E}[acc_{\textrm{past}}] = \mathbf{E}[acc_{\text{present}}], \quad \\ 
                   & H_1^\prime: \mathbf{E}[acc_{\text{past}}] > \mathbf{E}[acc_{\text{present}}]
    \end{aligned}
\end{equation}


Conversely, this hypothesis tests if there is a significant increase in model accuracy for past questions, indicating a preference for older or historical content. The p-value \( \widehat{p}^\prime \) quantifies the strength of evidence against the null hypothesis.
\end{hypo}

\paragraph{P-Value Computation} The p-values for assessing the shifts in accuracy are computed using the standard normal cumulative distribution function $\Phi$ as follows:
\begin{equation} \small \label{eq:pvalue}
        \widehat{p} = 1 - \Phi\left(\frac{acc_{\text{1}} - acc_{\text{2}}}{\sqrt{\frac{acc_{\text{1}}(1 - acc_{\text{1}})}{n_{1}} + \frac{acc_{\text{2}}(1 - acc_{\text{2}})}{n_{2}}}}\right),
\end{equation}

 $acc_{\text{1}}$ and $acc_{\text{1}}$ could be  either $acc_{\textrm{past}}$ or $acc_{\textrm{present}}$, which is  based on which Hypothesis Testing is used. Note that the two Hypothesis Testings are symmetrical.

Detailed experiment settings are listed in Appendix \ref{app:exp1}.

\begin{table*}[t]
\caption{Comparison of Model Performance in Different Pre-Release Periods (Compared to 0-20 months before Release), uses asterisks to denote significance levels:$^{\dagger}$ (p < 0.05), $^{\dagger\dagger }$(p < 0.01), and $^{\dagger\dagger\dagger }$ (p < 0.001), \textcolor{darkgreen}{$\pmb{\pmb{\pmb{\pmb{\rightarrow}}}}$}  refers to \textcolor{darkgreen}{Neophilia} Bias, \textcolor{navy}{$\pmb{\pmb{\pmb{\leftarrow}}}$}  refers to \textcolor{navy}{Nostalgia} Bias. 
}
\label{table:acc_skewness_bias}
\centering
\footnotesize
\addtolength\tabcolsep{-2.7pt}
\begin{tabular}{lrrr}
\toprule
Models & \parbox{30mm} {\centering 20-40 Months \\ Before Release} & \parbox{30mm} {\centering 40-60 Months \\ Before Release} & \parbox{30mm} {\centering 60-80 Months \\ Before Release}   \\
\midrule
OPT-13B & \textcolor{maroon}{\pmb{\pmb{\pmb{$-$}}}}~~~~~~~~~~~~~~~~~ & \textcolor{navy}{$\pmb{\pmb{\pmb{\leftarrow}}}$} $^{\dagger\dagger\dagger }$~~~~~~~~~~~ & \textcolor{navy}{$\pmb{\pmb{\pmb{\leftarrow}}}$} $^{\dagger\dagger }$ ~~~~~~~~~~~~~\\
OPT-2.7B & \textcolor{maroon}{\pmb{\pmb{\pmb{$-$}}}}~~~~~~~~~~~~~~~~~ & \textcolor{darkgreen}{$\pmb{\pmb{\pmb{\pmb{\rightarrow}}}}$}~$^{\dagger}$~~~~~~~~~~~~~~~& \textcolor{maroon}{\pmb{\pmb{\pmb{$-$}}}}~~~~~~~~~~~~~~~~~ \\
LLaMA-7B & \textcolor{maroon}{\pmb{\pmb{\pmb{$-$}}}}~~~~~~~~~~~~~~~~~ & \textcolor{navy}{$\pmb{\pmb{\pmb{\leftarrow}}}$} $^{\dagger\dagger\dagger }$~~~~~~~~~~~ & \textcolor{navy}{$\pmb{\pmb{\pmb{\leftarrow}}}$} $^{\dagger\dagger }$ ~~~~~~~~~~~~~\\
Pythia-12B & \textcolor{darkgreen}{$\pmb{\pmb{\pmb{\pmb{\rightarrow}}}}$} $^{\dagger\dagger }$ ~~~~~~~~~~~~& \textcolor{navy}{$\pmb{\pmb{\pmb{\leftarrow}}}$} $^{\dagger\dagger\dagger }$ ~~~~~~~~~~~& \textcolor{navy}{$\pmb{\pmb{\pmb{\leftarrow}}}$} $^{\dagger\dagger\dagger }$~~~~~~~~~~~ \\
Falcon-rw-1B & \textcolor{maroon}{\pmb{\pmb{\pmb{$-$}}}}~~~~~~~~~~~~~~~~~ & \textcolor{maroon}{\pmb{\pmb{\pmb{$-$}}}}~~~~~~~~~~~~~~~~~ & \textcolor{maroon}{\pmb{\pmb{\pmb{$-$}}}}~~~~~~~~~~~~~~~~~ \\
Baichuan-7B-Chat & \textcolor{maroon}{\pmb{\pmb{\pmb{$-$}}}}~~~~~~~~~~~~~~~~~ & \textcolor{maroon}{\pmb{\pmb{\pmb{$-$}}}}~~~~~~~~~~~~~~~~~ & \textcolor{navy}{$\pmb{\pmb{\pmb{\leftarrow}}}$}$~^{\dagger}$~~~~~~~~~~~~~~~\\
LLaMA-2-13B & \textcolor{maroon}{\pmb{\pmb{\pmb{$-$}}}}~~~~~~~~~~~~~~~~~ & \textcolor{navy}{$\pmb{\pmb{\pmb{\leftarrow}}}$} $^{\dagger\dagger\dagger }$~~~~~~~~~~~ & \textcolor{navy}{$\pmb{\pmb{\pmb{\leftarrow}}}$} $^{\dagger\dagger\dagger }$~~~~~~~~~~~ \\
LLaMA-2-7B & \textcolor{navy}{$\pmb{\pmb{\pmb{\leftarrow}}}$} $^{\dagger\dagger\dagger }$ ~~~~~~~~~~ & \textcolor{navy}{$\pmb{\pmb{\pmb{\leftarrow}}}$}~$^{\dagger}$~~~~~~~~~~~~~~~& \textcolor{maroon}{\pmb{\pmb{\pmb{$-$}}}}~~~~~~~~~~~~~~~~~ \\
LLaMA-2-7B-Chat & \textcolor{maroon}{\pmb{\pmb{\pmb{$-$}}}}~~~~~~~~~~~~~~~~~ & \textcolor{navy}{$\pmb{\pmb{\pmb{\leftarrow}}}$} $^{\dagger\dagger\dagger }$~~~~~~~~~~~ & \textcolor{navy}{$\pmb{\pmb{\pmb{\leftarrow}}}$} $^{\dagger\dagger\dagger }$~~~~~~~~~~~ \\
Zhongjing-Base & \textcolor{maroon}{\pmb{\pmb{\pmb{$-$}}}}~~~~~~~~~~~~~~~~~ & \textcolor{maroon}{\pmb{\pmb{\pmb{$-$}}}}~~~~~~~~~~~~~~~~~ & \textcolor{navy}{$\pmb{\pmb{\pmb{\leftarrow}}}$} $^{\dagger\dagger\dagger }$~~~~~~~~~~~ \\
InternLM-Chat-7B & \textcolor{navy}{$\pmb{\pmb{\pmb{\leftarrow}}}$} $^{\dagger\dagger }$~~~~~~~~~~~~~ & \textcolor{maroon}{\pmb{\pmb{\pmb{$-$}}}}~~~~~~~~~~~~~~~~~ & \textcolor{maroon}{\pmb{\pmb{\pmb{$-$}}}}~~~~~~~~~~~~~~~~~ \\
Baichuan2-7B-Chat & \textcolor{navy}{$\pmb{\pmb{\pmb{\leftarrow}}}$} $^{\dagger\dagger }$ ~~~~~~~~~~~~~& \textcolor{maroon}{\pmb{\pmb{\pmb{$-$}}}}~~~~~~~~~~~~~~~~~ & \textcolor{maroon}{\pmb{\pmb{\pmb{$-$}}}}~~~~~~~~~~~~~~~~~ \\
Mistral-7B-v0.1 & \textcolor{darkgreen}{$\pmb{\pmb{\pmb{\pmb{\pmb{\rightarrow}}}}}$}~$^{\dagger}$~~~~~~~~~~~~~~ & \textcolor{navy}{$\pmb{\pmb{\pmb{\leftarrow}}}$} $^{\dagger\dagger }$ ~~~~~~~~~~~~~& \textcolor{navy}{$\pmb{\pmb{\pmb{\leftarrow}}}$} $^{\dagger\dagger\dagger }$~~~~~~~~~~~ \\
Phi-1.5 & \textcolor{navy}{$\pmb{\pmb{\pmb{\leftarrow}}}$}~$^{\dagger}$~~~~~~~~~~~~~~~& \textcolor{navy}{$\pmb{\pmb{\pmb{\leftarrow}}}$} $^{\dagger\dagger }$ ~~~~~~~~~~~~~& \textcolor{navy}{$\pmb{\pmb{\pmb{\leftarrow}}}$} $^{\dagger\dagger\dagger }$~~~~~~~~~~~ \\
Baichuan2-13B-Base & \textcolor{navy}{$\pmb{\pmb{\pmb{\leftarrow}}}$} $^{\dagger\dagger }$ ~~~~~~~~~~~~~& \textcolor{navy}{$\pmb{\pmb{\pmb{\leftarrow}}}$} $^{\dagger\dagger\dagger }$~~~~~~~~~~~ & \textcolor{navy}{$\pmb{\pmb{\pmb{\leftarrow}}}$} $^{\dagger\dagger\dagger }$~~~~~~~~~~~ \\
Baichuan2-13B-Chat & \textcolor{navy}{$\pmb{\pmb{\pmb{\leftarrow}}}$}~$^{\dagger}$~~~~~~~~~~~~~~~& \textcolor{navy}{$\pmb{\pmb{\pmb{\leftarrow}}}$} $^{\dagger\dagger\dagger }$~~~~~~~~~~~ & \textcolor{navy}{$\pmb{\pmb{\pmb{\leftarrow}}}$} $^{\dagger\dagger\dagger }$~~~~~~~~~~~ \\
Colossal-LLaMA-2-7B-Base & \textcolor{navy}{$\pmb{\pmb{\pmb{\leftarrow}}}$} $^{\dagger\dagger }$ ~~~~~~~~~~~~~& \textcolor{maroon}{\pmb{\pmb{\pmb{$-$}}}}~~~~~~~~~~~~~~~~~ & \textcolor{navy}{$\pmb{\pmb{\pmb{\leftarrow}}}$}~$^{\dagger}$~~~~~~~~~~~~~~~\\
Qwen-14B-Chat & \textcolor{navy}{$\pmb{\pmb{\pmb{\leftarrow}}}$} $^{\dagger\dagger }$ ~~~~~~~~~~~~~& \textcolor{navy}{$\pmb{\pmb{\pmb{\leftarrow}}}$} $^{\dagger\dagger }$ ~~~~~~~~~~~~~& \textcolor{navy}{$\pmb{\pmb{\pmb{\leftarrow}}}$} $^{\dagger\dagger\dagger }$~~~~~~~~~~~ \\
Qwen-7B & \textcolor{maroon}{\pmb{\pmb{\pmb{$-$}}}}~~~~~~~~~~~~~~~~~ & \textcolor{maroon}{\pmb{\pmb{\pmb{$-$}}}}~~~~~~~~~~~~~~~~~ & \textcolor{maroon}{\pmb{\pmb{\pmb{$-$}}}}~~~~~~~~~~~~~~~~~ \\
Skywork-13B-Base & \textcolor{maroon}{\pmb{\pmb{\pmb{$-$}}}}~~~~~~~~~~~~~~~~~ & \textcolor{maroon}{\pmb{\pmb{\pmb{$-$}}}}~~~~~~~~~~~~~~~~~ & \textcolor{maroon}{\pmb{\pmb{\pmb{$-$}}}}~~~~~~~~~~~~~~~~~ \\
Zephyr-7B-beta & \textcolor{navy}{$\pmb{\pmb{\pmb{\leftarrow}}}$}~$^{\dagger}$~~~~~~~~~~~~~~~& \textcolor{navy}{$\pmb{\pmb{\pmb{\leftarrow}}}$} $^{\dagger\dagger\dagger }$~~~~~~~~~~~ & \textcolor{navy}{$\pmb{\pmb{\pmb{\leftarrow}}}$} $^{\dagger\dagger\dagger }$~~~~~~~~~~~ \\
Yi-6B & \textcolor{maroon}{\pmb{\pmb{\pmb{$-$}}}}~~~~~~~~~~~~~~~~~ & \textcolor{maroon}{\pmb{\pmb{\pmb{$-$}}}}~~~~~~~~~~~~~~~~~ & \textcolor{maroon}{\pmb{\pmb{\pmb{$-$}}}}~~~~~~~~~~~~~~~~~ \\
Yi-6B-Chat & \textcolor{navy}{$\pmb{\pmb{\pmb{\leftarrow}}}$}~$^{\dagger}$~~~~~~~~~~~~~~~& \textcolor{navy}{$\pmb{\pmb{\pmb{\leftarrow}}}$}~$^{\dagger}$~~~~~~~~~~~~~~~& \textcolor{navy}{$\pmb{\pmb{\pmb{\leftarrow}}}$}~$^{\dagger}$~~~~~~~~~~~~~~~\\
Qwen-1.8B & \textcolor{maroon}{\pmb{\pmb{\pmb{$-$}}}}~~~~~~~~~~~~~~~~~ & \textcolor{maroon}{\pmb{\pmb{\pmb{$-$}}}}~~~~~~~~~~~~~~~~~ & \textcolor{navy}{$\pmb{\pmb{\pmb{\leftarrow}}}$} $^{\dagger\dagger }$ ~~~~~~~~~~~~\\
RWKV-v5-Eagle-7B & \textcolor{maroon}{\pmb{\pmb{\pmb{$-$}}}}~~~~~~~~~~~~~~~~~ & \textcolor{maroon}{\pmb{\pmb{\pmb{$-$}}}}~~~~~~~~~~~~~~~~~ & \textcolor{maroon}{\pmb{\pmb{\pmb{$-$}}}}~~~~~~~~~~~~~~~~~ \\
Phi-2 & \textcolor{navy}{$\pmb{\pmb{\pmb{\leftarrow}}}$} $^{\dagger\dagger }$ ~~~~~~~~~~~~~& \textcolor{maroon}{\pmb{\pmb{\pmb{$-$}}}}~~~~~~~~~~~~~~~~~ & \textcolor{navy}{$\pmb{\pmb{\pmb{\leftarrow}}}$}~$^{\dagger}$~~~~~~~~~~~~~~~\\
Command R+ & \textcolor{navy}{$\pmb{\pmb{\pmb{\leftarrow}}}$} $^{\dagger\dagger }$~~~~~~~~~~~~~ & \textcolor{navy}{$\pmb{\pmb{\pmb{\leftarrow}}}$} $^{\dagger\dagger\dagger }$~~~~~~~~~~~ & \textcolor{navy}{$\pmb{\pmb{\pmb{\leftarrow}}}$} $^{\dagger\dagger\dagger }$~~~~~~~~~~~ \\
Mixtral-8x22B-Instruct-v0.1 & \textcolor{maroon}{\pmb{\pmb{\pmb{$-$}}}}~~~~~~~~~~~~~~~~~ & \textcolor{navy}{$\pmb{\pmb{\pmb{\leftarrow}}}$} $^{\dagger\dagger\dagger }$~~~~~~~~~~~ & \textcolor{navy}{$\pmb{\pmb{\pmb{\leftarrow}}}$} $^{\dagger\dagger\dagger }$~~~~~~~~~~~ \\
Phi-3-mini-4k-instruct & \textcolor{maroon}{\pmb{\pmb{\pmb{$-$}}}}~~~~~~~~~~~~~~~~~ & \textcolor{navy}{$\pmb{\pmb{\pmb{\leftarrow}}}$} $^{\dagger\dagger\dagger }$~~~~~~~~~~~ & \textcolor{navy}{$\pmb{\pmb{\pmb{\leftarrow}}}$} $^{\dagger\dagger }$ ~~~~~~~~~~~~\\
Qwen1.5-110B-Chat & \textcolor{navy}{$\pmb{\pmb{\pmb{\leftarrow}}}$}~$^{\dagger}$~~~~~~~~~~~~~~~& \textcolor{navy}{$\pmb{\pmb{\pmb{\leftarrow}}}$} $^{\dagger\dagger\dagger }$~~~~~~~~~~~ & \textcolor{navy}{$\pmb{\pmb{\pmb{\leftarrow}}}$} $^{\dagger\dagger\dagger }$~~~~~~~~~~~ \\
\bottomrule
\end{tabular}
\end{table*}

\vspace{4mm}

\subsubsection{Observation}

\textbf{Models tend to be stronger in earlier periods before release, with performance gradually declining over time.} Using the -20 to 0 months period as an anchor, if we assign the three periods to represent the near past (20 to 40 months before release), the middle past (40 to 60 months before release), and the distant past (60 to 80 months before release), we observe that the further back in time we go, the more pronounced the \textcolor{navy}{Nostalgia} Bias becomes, and it is also more significant. This suggests that the models gradually degrade during the pre-release period, rather than experiencing a rapid decline around the cutoff period.

\textbf{\textcolor{navy} {Nostalgia} Bias is prevalent.}As shown in Tab.~\ref{table:acc_skewness_bias}, among the biases, \textcolor{navy}{Nostalgia} is the most common, followed by periods where neither bias is significant (shown as \textcolor{maroon}{-}), with \textcolor{darkgreen}{Neophilia} Bias being the least frequent. 
Phi-1.5 and Baichuan2-13B series show \textcolor{navy}{Nostalgia} Bias in all three periods before release. 
Mistral-7B-v0.1 show \textcolor{darkgreen}{Neophilia} Bias in 20 to 40 months before release but show more significant \textcolor{navy}  {Nostalgia} in another two periods, which we attribute to random fluctuations rather than a consistent trend.

This observation reveals that models generally favor older knowledge and prefer information from more distant past periods. The distribution of knowledge is not uniform, nor does it tend to be closer to the present. This poses a potential risk in our utilization of these models because there is an expectation that the knowledge utilized should be as up-to-date as possible, ideally close to the model's release date.

\subsection{Temporal Degeneration}
\label{sec:Generalization}


\subsubsection{Protocol}

We discussed generally \textit{Temporal Degeneration} in \ref{sec:define_bias&degeneration}, here we need to define specifically.

\begin{definition}
\textbf{Temporal Degeneration}  The capability of future event prediction might be degraded over time due to the larger generalization gap brought by time.
\end{definition}

To qualitatively measure Temporal Degeneration, we propose a hypothesis testing like \ref{hyp_testing}.
\begin{hypo}
Temporal Degeneration~
\begin{equation} 
    \begin{aligned}
        T_3: \quad & H_0: \mathbf{E}[acc_{\text{future}}] = \mathbf{E}[acc_{\text{present}}], \\
                   & H_1: \mathbf{E}[acc_{\text{future}}] < \mathbf{E}[acc_{\text{present}}]
    \end{aligned}
\end{equation} 

Similar to Sec \ref{hyp_testing}, \( acc_{\text{future}} \) represents the model accuracy on questions proposed after the model's release. This hypothesis assesses whether there is a significant decline in model accuracy, indicating a degeneration over time. The hypothetical p-value is similar to that of Eq.~\ref{eq:pvalue}.
\end{hypo}

\begin{table*}[h!]
\small

\caption{Accuracy performance of various models across different periods. The presence of asterisks (*) denotes statistically significant decreases in performance during specific periods (p < 0.05 for *, p < 0.01 for **, and p < 0.001 for ***), benchmarked against their accuracy before release. Grey cells indicate the models have not been released in this period and represent the average scores of the models prior to their release.}

\label{table:accperformance}
\begin{center}
\renewcommand{\arraystretch}{1.1}
\addtolength\tabcolsep{-1.7pt}
\begin{tabular}{lllllllllll}
\toprule

\parbox{11mm}{Model} & \parbox{8mm}{23/1\\-23/2} & \parbox{8mm}{23/3\\-23/4} & \parbox{8mm}{23/5\\-23/6} & \parbox{8mm}{23/7\\-23/8} & \parbox{8mm}{23/9\\-23/10} & \parbox{8mm}{23/11\\-23/12} & \parbox{8mm}{24/1\\-24/2} & \parbox{8mm}{24/3\\-24/4} & \parbox{8mm}{24/5\\-24/6} & \parbox{8mm}{24/7\\-24/8} \\

\hline

Baichuan-13B-Chat           &\multicolumn{2}{c}{\cellcolor{gray!25}0.52}&                       0.26** &                       0.20** &                       0.36** &                       0.32** &                       0.33** &                       0.20** &                       0.23** &                       0.17** \\\
GPT-3.5-turbo-230613          &\multicolumn{2}{c}{\cellcolor{gray!25}0.49}&                       0.29** &                        0.37* &                         0.45 &                         0.44 &                        0.37* &                       0.29** &                       0.28** &                       0.26** \\\
GPT-4-230613                   &\multicolumn{2}{c}{\cellcolor{gray!25}0.6}&                       0.39** &                        0.43* &                       0.41** &                       0.38** &                         0.48 &                       0.40** &                       0.35** &                       0.36** \\\
Llama-2-13B              &\multicolumn{3}{c}{\cellcolor{gray!25}0.52}&                         0.43 &                         0.48 &                       0.36** &                        0.39* &                        0.40* &                       0.28** &                       0.34** \\\
Llama-2-7B               &\multicolumn{3}{c}{\cellcolor{gray!25}0.27}&                         0.26 &                         0.27 &                         0.22 &                         0.26 &                         0.33 &                       0.18** &                         0.24 \\\
LLaMA2-7B-Chat           &\multicolumn{3}{c}{\cellcolor{gray!25}0.45}&                        0.30* &                         0.46 &                       0.35** &                         0.37 &                         0.38 &                       0.27** &                       0.28** \\\
Baichuan2-13B-Base          &\multicolumn{3}{c}{\cellcolor{gray!25}0.49}&                       0.30** &                         0.46 &                        0.42* &                       0.28** &                        0.39* &                       0.27** &                       0.27** \\\
Baichuan2-13B-Chat          &\multicolumn{3}{c}{\cellcolor{gray!25}0.5}&                       0.33** &                         0.43 &                        0.41* &                       0.26** &                       0.29** &                       0.29** &                       0.33** \\\
Baichuan2-7B-Base           &\multicolumn{3}{c}{\cellcolor{gray!25}0.42}&                       0.26** &                        0.30* &                       0.24** &                         0.33 &                        0.33* &                       0.25** &                       0.22** \\\
Baichuan2-7B-Chat           &\multicolumn{3}{c}{\cellcolor{gray!25}0.37}&                         0.35 &                         0.43 &                         0.31 &                         0.37 &                         0.40 &                       0.23** &                        0.27* \\\
Colossal-LLaMA-2-7B-Base    &\multicolumn{4}{c}{\cellcolor{gray!25}0.42}&                         0.43 &                         0.37 &                        0.28* &                        0.33* &                       0.25** &                       0.26** \\\
Mistral-7B-v0.1             &\multicolumn{4}{c}{\cellcolor{gray!25}0.44}&                         0.41 &                        0.35* &                        0.28* &                        0.34* &                       0.25** &                       0.24** \\\
Phi-1.5                     &\multicolumn{4}{c}{\cellcolor{gray!25}0.41}&                         0.38 &                         0.38 &                         0.30 &                       0.26** &                        0.32* &                       0.27** \\\
Qwen-14B-Chat               &\multicolumn{4}{c}{\cellcolor{gray!25}0.39}&                       0.25** &                       0.24** &                         0.33 &                       0.27** &                       0.28** &                       0.19** \\\
Zephyr-7B-beta              &\multicolumn{4}{c}{\cellcolor{gray!25}0.4}&                         0.32 &                        0.32* &                        0.28* &                       0.28** &                        0.31* &                       0.19** \\\
Yi-6B                       &\multicolumn{4}{c}{\cellcolor{gray!25}0.29}&                         0.27 &                       0.16** &                         0.22 &                         0.33 &                        0.21* &                        0.21* \\\
Gemini                      &\multicolumn{5}{c}{\cellcolor{gray!25}0.43}&                       0.20** &                         0.37 &                       0.24** &                       0.21** &                       0.17** \\\
Qwen-1.8B                   &\multicolumn{5}{c}{\cellcolor{gray!25}0.35}&                       0.22** &                         0.37 &                         0.29 &                       0.20** &                       0.20** \\\
GPT-4-231106                   &\multicolumn{5}{c}{\cellcolor{gray!25}0.66}&                       0.45** &                       0.39** &                       0.36** &                       0.47** &                       0.43** \\\
Phi-2                       &\multicolumn{5}{c}{\cellcolor{gray!25}0.31}&                        0.23* &                         0.22 &                         0.29 &                       0.18** &                       0.15** \\\
Claude-3-opus-20240229      &\multicolumn{6}{c}{\cellcolor{gray!25}0.53}&                         0.52 &                       0.37** &                         0.47 &                       0.36** \\\
Claude-3-sonnet-20240229    &\multicolumn{6}{c}{\cellcolor{gray!25}0.23}&                         0.24 &                       0.13** &                       0.10** &                       0.07** \\\
Command R+              &\multicolumn{7}{c}{\cellcolor{gray!25}0.54}&                       0.37** &                        0.43* &                       0.33** \\\
DeepSeek-V2-Chat            &\multicolumn{7}{c}{\cellcolor{gray!25}0.58}&                       0.49** &                       0.46** &                       0.35** \\\
Mixtral-8x22B-Instruct-v0.1 &\multicolumn{7}{c}{\cellcolor{gray!25}0.56}&                       0.39** &                       0.43** &                       0.35** \\\
Phi-3-mini-4k-Instruct      &\multicolumn{7}{c}{\cellcolor{gray!25}0.38}&                       0.26** &                       0.24** &                       0.26** \\\
Qwen1.5-110B-Chat           &\multicolumn{7}{c}{\cellcolor{gray!25}0.56}&                       0.40** &                        0.45* &                       0.36** \\\
Claude-3.5-Sonnet-20240620  &\multicolumn{8}{c}{\cellcolor{gray!25}0.63}&                       0.48** &                       0.23** \\\
Gemini-1.5-Pro              &\multicolumn{8}{c}{\cellcolor{gray!25}0.5}&                         0.47 &                       0.35** \\\
Gemini-1.5-flash            &\multicolumn{8}{c}{\cellcolor{gray!25}0.39}&                       0.23** &                        0.22* \\\
Qwen2-72B-Instruct          &\multicolumn{8}{c}{\cellcolor{gray!25}0.6}&                       0.45** &                       0.40** \\\
LLaMA-3.1-405B-Instruct     &\multicolumn{9}{c}{\cellcolor{gray!25}0.5}&                         0.41 \\\
GPT-4o                      &\multicolumn{9}{c}{\cellcolor{gray!25}0.62}&                       0.39** \\\
GPT-4o-mini-2024-07-18      &\multicolumn{9}{c}{\cellcolor{gray!25}0.48}& 0.28** \\

\hline

\label{grey_table}
\end{tabular}
\end{center}
    \vspace{-7mm}
\end{table*}

\subsubsection{Observation}

\textbf{Earlier models perform worse, while newer advanced models perform better in future predictions.}
As the latest scores in Table \ref{grey_table} show, earlier models, show a significant performance decline across multiple time periods. For example, LLaMA-7B’s accuracy dropped from 0.11 in March 2023 to 0.08 by March 2024, with no improvement in later evaluation periods. In contrast, newer models exhibit much stronger future prediction capabilities.

\textbf{Close-source models perform well initially but struggle with temporal generalization, while open-source models show better stability.}
As is shown in Appendix \ref{app:full_comparision_open_close}, in both pre- and post-release scenarios, proprietary models such as GPT-4 and Claude-3.5 slightly outperform their open-source counterparts. However, when predicting outcomes for events known only post-release, all models exhibited substantial declines in performance. 

Our analysis highlights models with less than 31\% decline in post-release accuracy in \textbf{bold}, illustrating relatively stable performance. Conversely, models that experienced declines greater than 39\% are marked by \underline{underline}, indicating substantial performance degeneration. Notably, models such as Command R+, Mixtral-8x22B-Instruct-v0.1, DeepSeek-V2-Chat, and LLaMA-3.1-405B-Instruct demonstrate slower declines in accuracy post-release.  Conversely, models like Claude-3.5-Sonnet-20240620, GPT-4-231106, and GPT-4o show rapid declines in performance,

The case studies of model iterations within the Claude series and the GPT models illustrate the dynamics of model evolution and generalization capabilities over time. Notably, the Claude-3-opus-20240229 model shows the least decline, indicating a potentially better tuning for broader applications or general datasets. This contrasts with the more significant declines seen in Claude-3-sonnet-20240229 and Claude-3.5-Sonnet-20240620, suggesting that these iterations may have undergone specific architectural tweaks, utilized different training data scopes, or implemented optimization strategies that did not generalize as effectively.

Similarly, the progression from GPT-3.5-turbo-230613 to GPT-4-231106 shows an increasing trend in both pre-release accuracy and the magnitude of decline post-release, with the GPT-4-231106 model experiencing the largest drop (39.39\%). This trend could indicate a push towards optimizing these models for higher initial accuracy, possibly at the expense of their ability to generalize effectively over time.

\textbf{Miniature versions of models show significant declines in temporal performance.}
The data reveals that miniature versions of models underperform compared to their full-sized counterparts in temporal tasks. For instance, Gemini.pro had an accuracy of 0.37 in March 2024, but this dropped sharply to 0.17 by July 2024, indicating a significant loss in precision for temporal tasks. Similarly, the mini version of GPT-4 showed considerably lower performance across multiple time periods, particularly in April 2024, where its accuracy was just 0.21, around 0.1 lower than the full-sized GPT-4 in the same period. This illustrates that while mini models improve inference efficiency, they exhibit poor generalization over time, with performance rapidly declining in long-term prediction tasks.

\section{Conclusion and Community Call to Action}
We rigorously define and quantify temporal generalization and bias within LLMs, uncovering significant challenges in predicting and adapting to future contexts. Our experiments reveal that LLMs often struggle with temporal biases, affecting reliability in dynamic scenarios requiring accurate forecasting. The evaluation framework and datasets we introduce allow for an in-depth assessment of model performance over time. The findings highlight shortcomings in existing benchmarks related to temporal shifts and underscore the urgent need for improvements in LLM training and updating processes to enhance adaptability and mitigate biases.

\vspace{-1mm}

Specifically, we urge the community to take note of this phenomenon: models tend to better comprehend earlier events.
~\citet{cheng2024dated} suggests that models are most familiar with knowledge up to the year 2020. 
However, our findings indicate that the understanding of world events (even in cases where models can remember) lingers around 2015, with a trend towards even earlier years (beyond the time range of our test data). 
This observation is contrary to our initial hypothesis that models may overfit current events, which is significantly more concerning. It suggests that caution is needed, especially for those hoping to use model judgments directly.

\newpage
\newpage

\section*{Acknowledgements}
This work was supported by the Shenzhen Science and Technology Program (JCYJ20220818103001002), Shenzhen Doctoral Startup Funding (RCBS20221008093330065), Tianyuan Fund for Mathematics of National Natural Science Foundation of China (NSFC) (12326608), Shenzhen Key Laboratory of Cross-Modal Cognitive Computing (grant number ZDSYS20230626091302006), and Shenzhen Stability Science Program 2023.

\section*{Future Work}

Moving forward, it is crucial to refine LLM development and evaluation techniques to better handle evolving informational environments, ensuring their continued relevance and utility in various applications.

Looking ahead, we suggest future work includes several ambitious goals aimed at enhancing the versatility and depth of our evaluation framework. Key among these objectives are:
\begin{itemize}
    \item Exploring mitigation strategies, through we conducted preliminary investigations into prompt engineering strategies that could potentially mitigate temporal performance degradation, see Abstract \ref{preliminary_Mitigation_Strategies}. 
    \item Integrating a wider array of models, particularly focusing on those based on the Transformer architecture, such as Griffin and Mamba models.
    \item Expanding the number of checkpoints accessed for a more comprehensive temporal analysis of model performance.
    \item Delving into the relationship between token-level loss metrics and overall scores, which could offer more granular insights into model capabilities.
    \item Developing methodologies to assess the obsolescence of benchmarks and models, potentially offering a dynamic approach to evaluating the relevance and efficiency of language models over time.
\end{itemize}
These initiatives represent our commitment to advancing the state of the art in language model evaluation, ensuring that our tools remain at the forefront of technological progress and innovation.

\section*{Limitations}
\label{sec_limitation}

While we strive to collect textual data from a variety of sources such as news articles, scientific papers, and forum discussions, the proportion and diversity of the texts may still fall short of fully reflecting the diversity and complexity of language. Also, the English-centric nature of the questions and answers limits the benchmark's applicability to non-English speaking regions and global events. Moreover, such evaluations on text loss require access to the model's logits, which is challenging for closed-source models or models accessed via APIs.

\section*{Potential Risks}
\label{sec_risk}
Bias and Fairness Risks: When used to make predictions about future events, language models can inadvertently perpetuate or amplify biases present in their training data. This could lead to biased forecasts that reflect and potentially exacerbate existing societal prejudices. Such biased outputs may result in models that are unfair or discriminatory, which could harm individuals or groups and damage the reputation of the organizations involved.

\section*{Ethics Statement}
\label{Ethics_Statement}
We fully acknowledge the need to adhere to ethical standards in handling and generating textual data. Hence, we commit to:
\begin{itemize}

    \item We utilize various datasets in our experiments, including publicly available sources like news articles and academic papers, which do not contain personally identifiable information or offensive content. However, some datasets created or used during our study contain sensitive information or are bound by specific usage agreements that prevent public sharing. Therefore, these datasets are utilized strictly within our experiments and are not released as part of our open-source materials, adhering to ethical guidelines and agreements.

    \item \textbf{Responsibility for Content Generated by Open-Source Models:} We recognize that content generated by our system or through the use of open-source models involved in the evaluation might reflect or amplify biases. While efforts are made to minimize these biases and improve the fairness of the content, we cannot take responsibility for the generated content. We encourage users and developers to critically engage with these models and remain vigilant of their outputs.
    \item \textbf{Addressing Data Leak Issues:} Our system aims to tackle data leak issues by dynamically generating benchmarks and ensuring models generalize to new or future texts. We commit to continuously monitoring and updating our evaluation methods to avoid potential data leakage, ensuring the validity and reliability of our assessment results.
    \item \textbf{Transparency:} We are committed to maintaining the highest level of transparency in our work, including our evaluation methods, data sources used, and assessment outcomes. We encourage open critique and suggestions for improvement of our methodology, aiming to foster knowledge sharing and technological advancement.
\end{itemize}

We believe that through this approach, we can offer a valuable resource to the research and development community, while fostering deep reflection and discussion on ethical issues in the artificial intelligence field.

\section*{Licence}
\label{Licence}

\subsection*{Good Judgment Inc.}
For personal use of its services and content, a limited, revocable, non-exclusive license is granted. Without explicit written permission from Good Judgment Inc., this license restricts the collection, aggregation, copying, duplication, display, or derivative use of its content.

\subsection*{arXiv}
The arXiv dataset is licensed under the Creative Commons Zero (CC0 1.0) license. This means it is in the public domain, and users can freely use, modify, and distribute the data without any restrictions.

\subsection*{Wikipedia}
Wikipedia content is licensed under the Creative Commons Attribution-ShareAlike (CC BY-SA) license. This allows users to share and adapt the content as long as appropriate credit is given, and any derived works are distributed under the same license.

Other text will not be released.

\bibliography{anthology,paper}
\bibliographystyle{acl_natbib}

\appendix
\newpage

\section{Related Work}

Given the limitations of traditional methods, researchers are exploring new ways to measure model performance more comprehensively. Recently, several methods have been proposed to evaluate models at different times and in various contexts~\citet{fatemi2024tot, kasai2024realtimeqa, vu2023freshqa}. TimeQA~\citet{chen2021timeqa} evaluates a model's ability to handle temporal evolution, while Situated QA~\citet{zhang21situatedqa} focuses on a model's performance in specific contexts. StreamingQA~\citet{liška2022streamingqa} introduces the concept of continuous information flow, while RealTimeQA~\citet{kasai2024realtimeqa} and FreshQA~\citet{vu2023freshqa} emphasize real-time and up-to-date information processing capabilities. 
LatestEval ~\citet{LatestEval} utilizes the most recent text to construct questions and answers, covering six most frequently queried categories including explanation, summary, reason, demonstrations, existence, and possible usage.

However, it's essential to clarify that RealTimeQA~\citet{kasai2024realtimeqa} and FreshQA~\citet{vu2023freshqa}, which continuously update topics, are inclined towards knowledge. For example, RealTimeQA~\citet{kasai2024realtimeqa} has such a sample in their paper, "Which wildly popular show was recently greenlit for a new season?" (Answer: Squid Game). These methods are typically used for evaluating retrieval-augmented generation (rag) series models. On the other hand, methods like LatestEval ~\citet{LatestEval}, which construct questions using the most recent information, lean towards enhancing reading comprehension.

In contrast to these approaches, our focus is on testing the model's understanding of the world, as demonstrated through prediction. Our method of evaluation is based on facts and computation loss. This ensures that our evaluation is resistant to data contamination, objective, and free from biases in question construction and judgment. This approach allows us to examine the model's ability to process and predict real-world events accurately and efficiently.

\paragraph{Future-knowledge Prediction}
Previous research has focused on how to automatically predict real-world events. For instance, \citet{zou2022forecasting} constructed a dataset of questions obtained from tournament predictions as an evaluation benchmark for automatic forecasting. However, this dataset may already be outdated because its knowledge is likely already embedded within current LLMs. Recently, studies have shown that LLMs can match human performance in the task of forecasting the future. For example, \citet{Sch2024Wisdom} demonstrates that the predictive performance of an ensemble of multiple LLMs is superior to that of a single LLM, and it even surpasses the performance of a group of human forecasters.

\textbf{Model selection and Experimental Setting}
\label{app:exp2}
We selected a set of models due to their diversity and significance in advancing the capabilities of NLP systems. They are not only benchmarks of their respective sizes, ranging from 1 billion to 40 billion parameters, but also exemplify the rapid progression in NLP technology.
We set the context size to 2048, then compute the sum of the negative logarithms of each segment. We then divide the computed sum by the length of the text under UTF-8 encoding. Throughout this experiment, approximately 300 GPU hours on NVIDIA A800 were consumed.

\section{Exploring Mitigation Strategies} \label{preliminary_Mitigation_Strategies}

While our primary focus is temporal evaluation rather than solution development, we conducted preliminary investigations into prompt engineering strategies that could potentially mitigate temporal performance degradation. 

\subsection{Try Different Prompt to Mitigate} Motivated by observed safety constraints and refusal patterns in base models and the result of \citet{pham2024basechatgptusedforecasting}, we designed five specialized prompts to investigate how different framing strategies affect temporal prediction:

\begin{figure*}[!ht]
    \centering
    \begin{minipage}{0.48\textwidth}
        \includegraphics[width=\textwidth ]{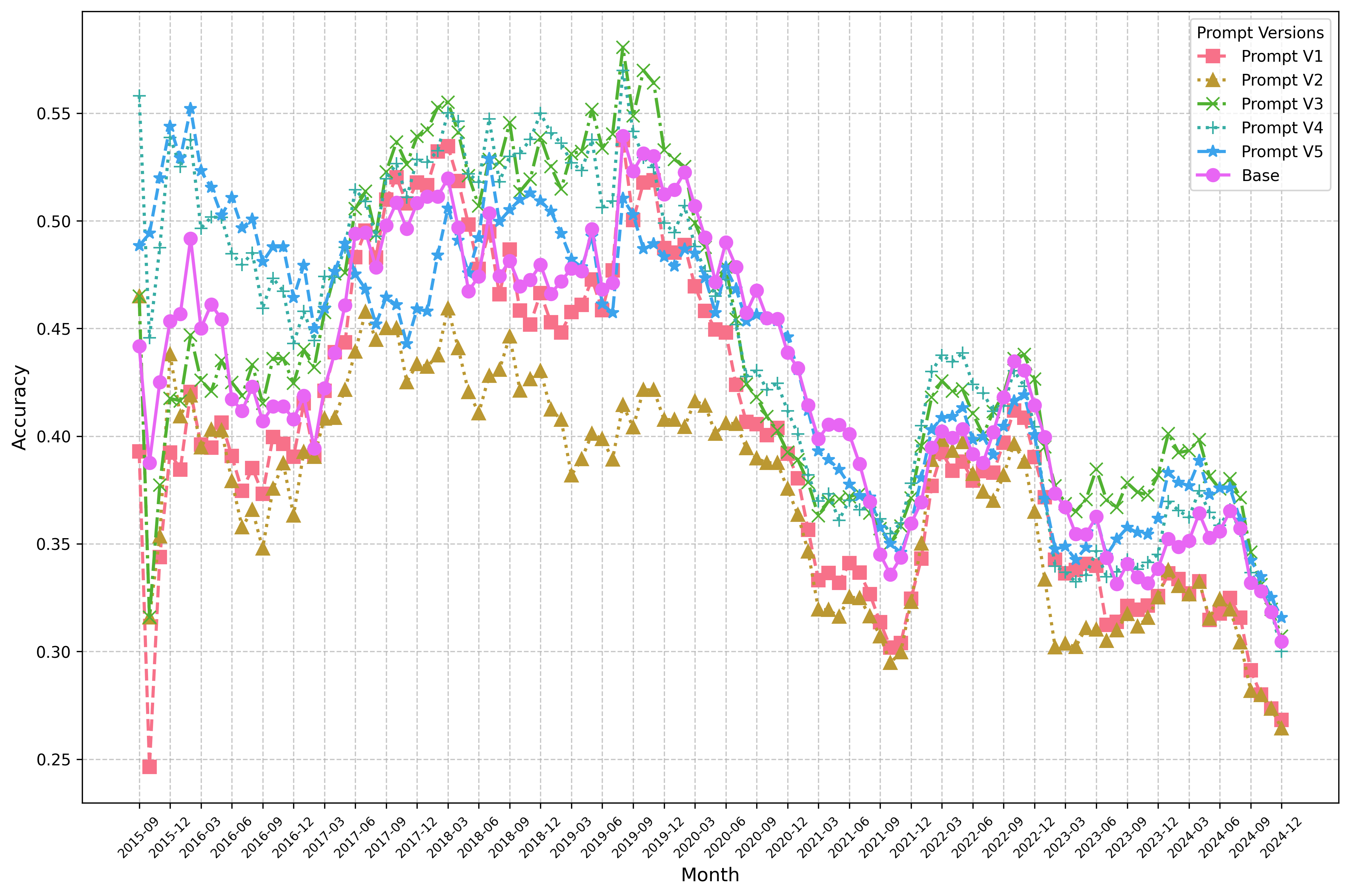}
        \caption{Accuracy of LlaMA3.1-8B-Instruct with different prompts}
        \label{fig:LLaMA3.1-8B-Instruct_accuracy}
    \end{minipage}\hfill
    \begin{minipage}{0.48\textwidth}
        \includegraphics[width=\textwidth]{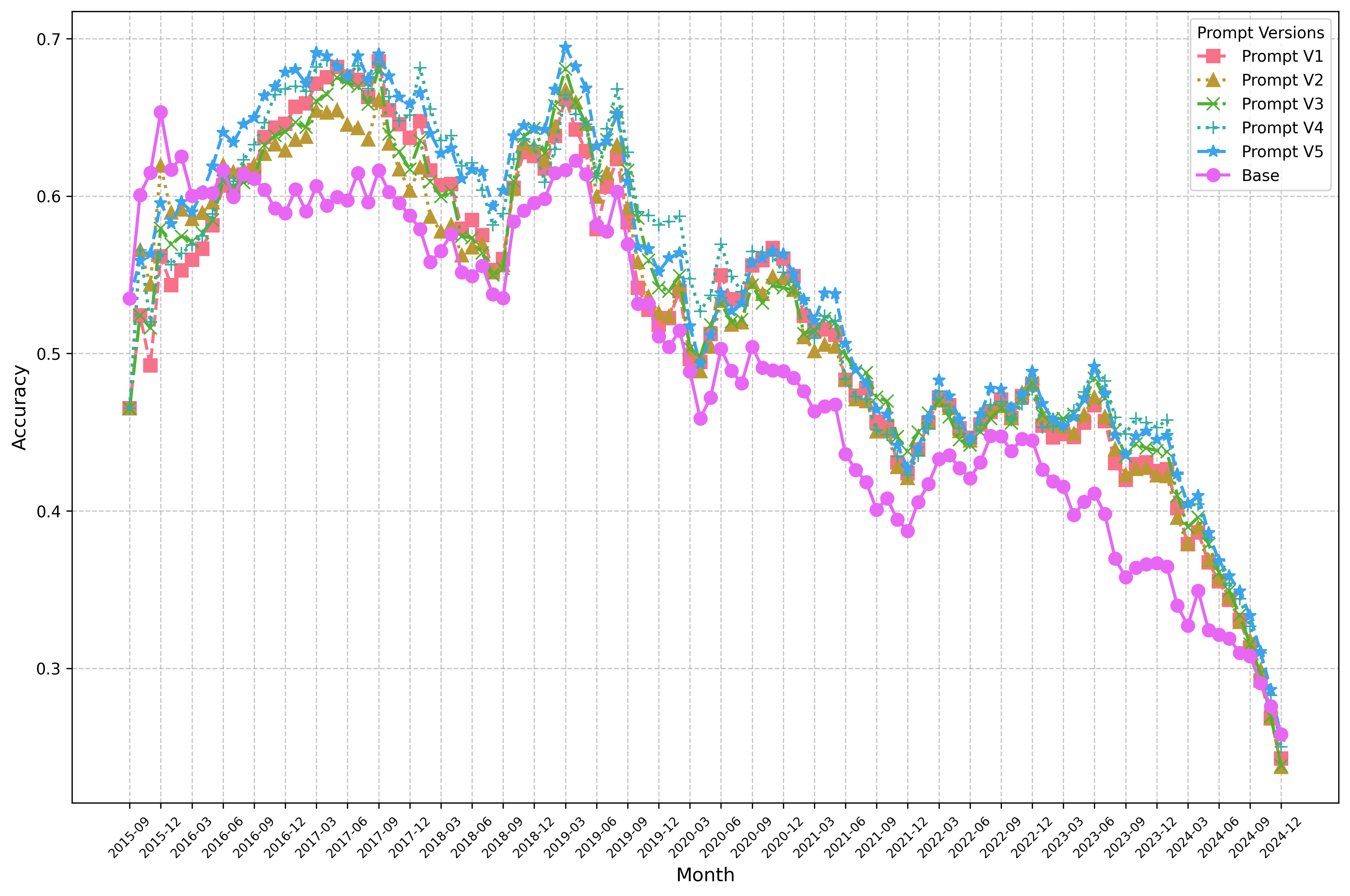}
        \caption{Accuracy of Qwen2.5-7B with different prompts}
        \label{fig:Qwen2.5-7B_accuracy}
    \end{minipage}
    \vspace{5mm} 
    \begin{minipage}{0.48\textwidth}
        \includegraphics[width=\textwidth]{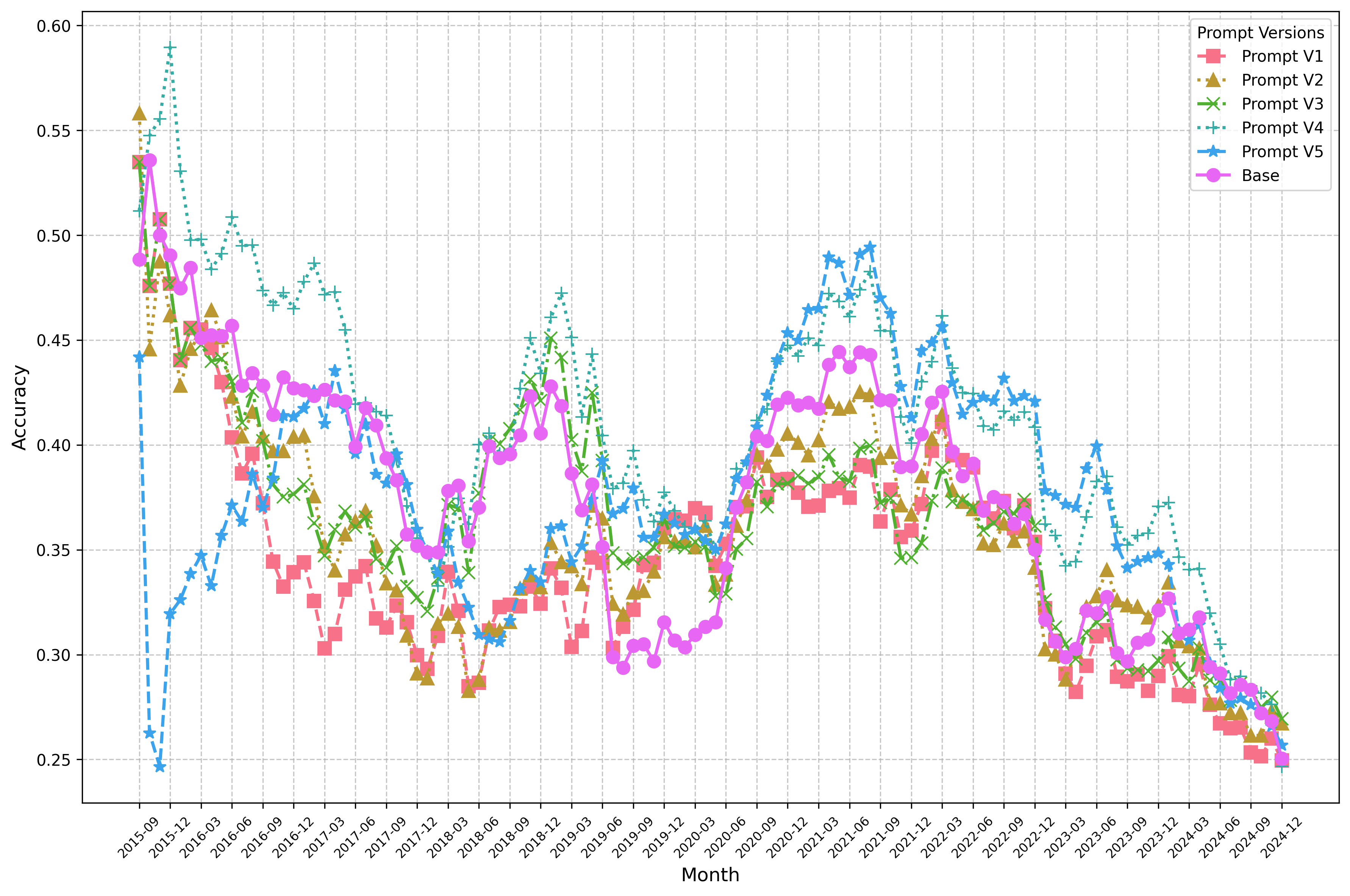}
        \caption{Accuracy of Baichuan2-7B-Chat with different prompts}
        \label{fig:Baichuan2-7B-Chat_accuracy}
    \end{minipage}\hfill
    \begin{minipage}{0.48\textwidth}
        \includegraphics[width=\textwidth]{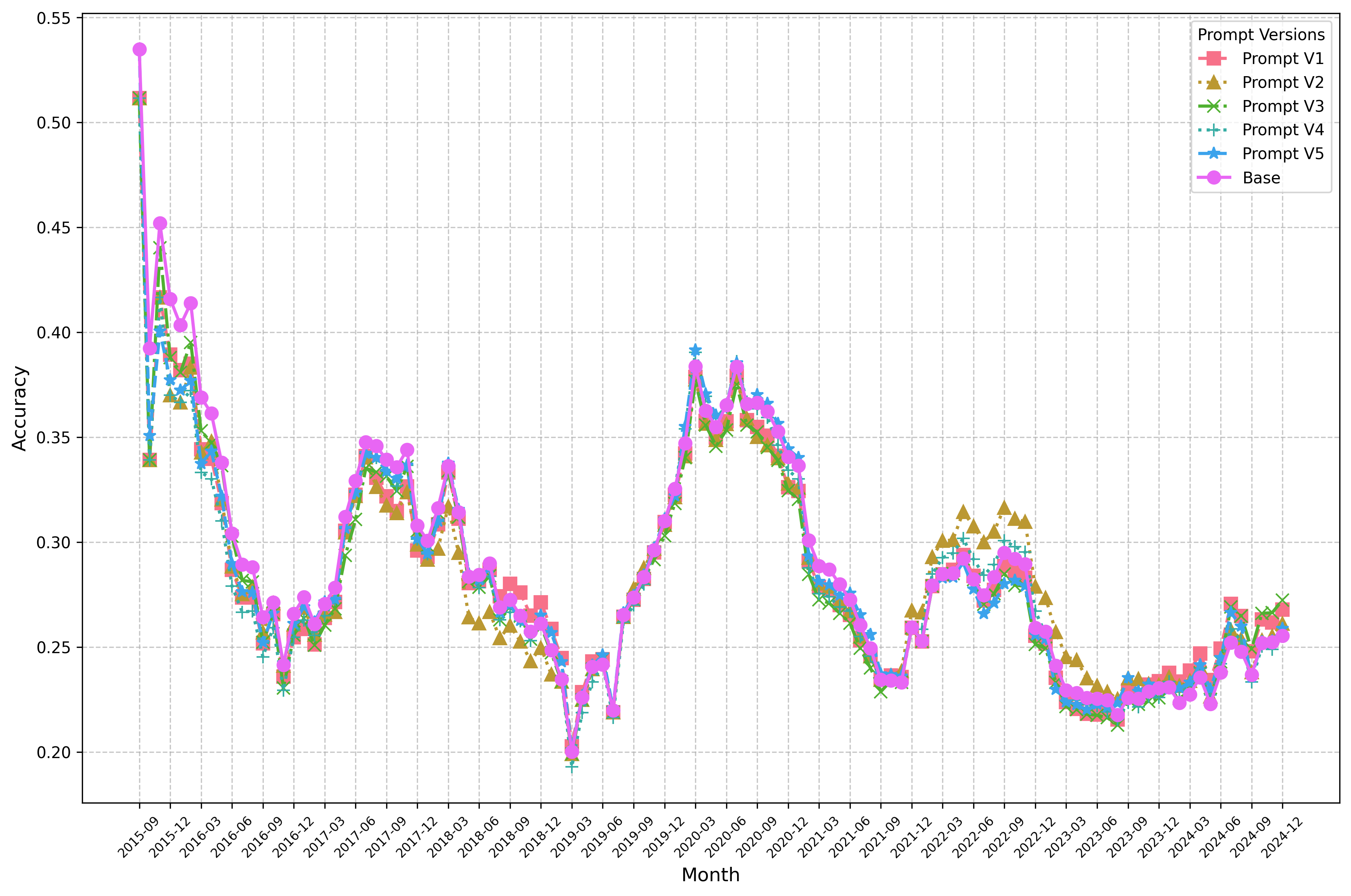}
        \caption{Accuracy of Yi-6B with different prompts}
        \label{fig:Yi-6B_accuracy}
    \end{minipage}
    \vspace{-2mm}
\end{figure*}

\vspace{-3mm}

\begin{itemize} 
    \item \textbf{Base}: Used direct incentivization ("tip of \$200") and explicit guessing encouragement to counter safety-driven refusals.
    \vspace{-3mm}
    \item \textbf{Dream Framework (v1)}: Encourages speculative thinking by positioning the model as a visionary oracle.
    \vspace{-3mm}
    \item \textbf{Time Travel (v2)}: Frames predictions as historical memories from a future perspective.
    \vspace{-3mm}
    \item \textbf{Multiverse (v3)}: Leverages parallel universe observations to bypass temporal constraints.
    \vspace{-3mm}
    \item \textbf{Precise Thinking (v4)}: Emphasizes structured temporal reasoning chains.
    \vspace{-3mm}
    \item \textbf{Future Recall (v5)}: Simulates retrospective analysis from a future vantage point.
    \vspace{-3mm}
\end{itemize}

See Figure \ref{fig:LLaMA3.1-8B-Instruct_accuracy}, \ref{fig:Qwen2.5-7B_accuracy}, \ref{fig:Baichuan2-7B-Chat_accuracy} and \ref{fig:Yi-6B_accuracy}. Specifically, different models respond variably to the designed prompts. For Yi-6B, there were scarcely any discernible differences in performance across the several prompts we employed. In contrast, for the Qwen2.5-7B model, all modified prompts generally outperformed the base prompt in most scenarios.
Analyzing the results from Llama3.1-8B-Instruct, Qwen2.5-7B and Baichuan2-7B-Chat, we found that prompts v5({Future Recall) and v4(Precise Thinking) demonstrated relatively better effectiveness. The specific impacts vary from model to model, which is closely related to the model's instruction-following capabilities.

In addition, our exploratory analysis revealed significant performance variations across question categories, with the model achieving lowest accuracy on "Finance" (58.1\%), "Business" (55.4\%).  and "Economic Indicators" (62.3\%) categories, compared to stronger performance on "Foreign Policy", (81.2\%), "Security and Conflict" (79.6\%) and "Leader Entry/Exit" (83.7\%) categories.

It is plausible that fine-tuning models with analysis and prediction data (especially in these weak areas) could fully unlock their inner predictive potential.

\section{Full Table of Decline in Prediction Accuracy}
\label{app:full_comparision_open_close}
See Table \ref{table:full_comparision_open_close}.

\begin{table*}[h!]
\centering
\caption{Comparison of Powerful Open-source and Closed-source LLMs}
\label{table:full_comparision_open_close}
\begin{tabular}{|p{5cm}|p{1.5cm}|p{1.5cm}|p{1.5cm}|p{1.5cm}|}
\toprule
  Model &  Pre-release Accuracy &  First Post-release Accuracy &  Decline & Percentage Decline \\
\midrule
\underline{Baichuan-13B-Chat} &                  0.52 &                         0.23 &     0.29 &             55.77\% \\
GPT-3.5-turbo-230613 &                  0.49 &                         0.33 &     0.16 &             32.65\% \\
GPT-4-230613 &                  0.60 &                         0.41 &     0.19 &             31.67\% \\
\textbf{Llama-2-13B} &                  0.52 &                         0.45 &     0.07 &             12.50\% \\
\textbf{Llama-2-7B} &                  0.27 &                         0.27 &     0.01 &              1.85\% \\
\textbf{LLaMA2-7B-Chat} &                  0.45 &                         0.38 &     0.07 &             15.56\% \\
\textbf{Baichuan2-13B-Base} &                  0.49 &                         0.38 &     0.11 &             22.45\% \\
\textbf{Baichuan2-13B-Chat} &                  0.50 &                         0.38 &     0.12 &             24.00\% \\
Baichuan2-7B-Base &                  0.42 &                         0.28 &     0.14 &             33.33\% \\
\textbf{Baichuan2-7B-Chat} &                  0.37 &                         0.39 &    -0.02 &             -5.41\% \\
\textbf{Colossal-LLaMA-2-7B-Base} &                  0.42 &                         0.40 &     0.02 &              4.76\% \\
\textbf{Mistral-7B-v0.1} &                  0.44 &                         0.38 &     0.06 &             13.64\% \\
\textbf{Phi-1.5} &                  0.41 &                         0.38 &     0.03 &              7.32\% \\
Qwen-14B-Chat &                  0.39 &                         0.24 &     0.15 &             37.18\% \\
\textbf{Zephyr-7B-beta} &                  0.40 &                         0.32 &     0.08 &             20.00\% \\
Gemini &                  0.43 &                         0.29 &     0.14 &             33.72\% \\
\textbf{Yi-6B} &                  0.29 &                         0.22 &     0.07 &             25.86\% \\
\textbf{Qwen-1.8B} &                  0.35 &                         0.29 &     0.05 &             15.71\% \\
GPT-4-231106 &                  0.66 &                         0.42 &     0.24 &             36.36\% \\
\textbf{Phi-2} &                  0.31 &                         0.23 &     0.08 &             27.42\% \\
\textbf{Command R+} &                  0.54 &                         0.40 &     0.14 &             25.93\% \\
\textbf{Claude-3-opus-20240229} &                  0.53 &                         0.45 &     0.09 &             16.04\% \\
\textbf{Claude-3-sonnet-20240229} &                  0.23 &                         0.18 &     0.05 &             19.57\% \\
\textbf{DeepSeek-V2-Chat} &                  0.58 &                         0.47 &     0.10 &             18.10\% \\
\textbf{Mixtral-8x22B-Instruct-v0.1} &                  0.56 &                         0.41 &     0.15 &             26.79\% \\
Phi-3-mini-4k-Instruct &                  0.38 &                         0.25 &     0.13 &             34.21\% \\
\textbf{Qwen1.5-110B-Chat} &                  0.56 &                         0.43 &     0.14 &             24.11\% \\
\underline{Claude-3.5-Sonnet-20240620} &                  0.63 &                         0.35 &     0.28 &             43.65\% \\
\textbf{Gemini-1.5-Pro} &                  0.50 &                         0.41 &     0.09 &             18.00\% \\
\underline{Gemini-1.5-flash} &                  0.39 &                         0.23 &     0.17 &             42.31\% \\
\textbf{Qwen2-72B-Instruct} &                  0.60 &                         0.43 &     0.17 &             29.17\% \\
\textbf{LLaMA-3.1-405B-Instruct} &                  0.50 &                         0.41 &     0.09 &             18.00\% \\
GPT-4o &                  0.62 &                         0.39 &     0.23 &             37.10\% \\
\underline{GPT-4o-mini-2024-07-18} &                  0.48 &                         0.28 &     0.20 &             41.67\% \\
\bottomrule
\end{tabular}

\end{table*}

\section{Experimental setting}
\label{app:exp1}
For this experiment, we sourced all questions from \citet{GoodJudgmentOpen2023} using web scraping. The data collected includes initiation and conclusion dates, domains, questions, descriptions, and answers, which were appropriately ordered for further processing. Approximately 50 GPU hours on NVIDIA A800 were consumed during this experiment.

Up to 2024.10.1, we collected a total of 2769 questions from \citet{GoodJudgmentOpen2023}. The earliest question was proposed on 2015-09-01: "Will Congress pass a resolution disapproving the Joint Comprehensive Plan of Action?" The most recent question was closed on 2024-10-01: "What will be the 12-month percentage change in the US Consumer Price Index (CPI) for September 2024?"
Question samples and most common tags are in Appendix \ref{gjo_tag_count}.

\section{Testing Statistical Hypothesis} 
Due to the inherent randomness in data, directly comparing outcomes of two experiments is insufficient to justify the superiority of one outcome over the other. To address this issue, practitioners often resort to hypothesis testing to make convincing decisions, like those demonstrated in \citet{hao16hypo}. A board application of hypothesis testing lies in the A/B testing literature, which assesses the performance of various features of  a product and performs post-experiment inferences, as introduced in \citet{russac21abn}.   

\section{Details of Data Collection}
\label{app:collection}

The data for our system is sourced from a diverse set of online platforms, each selected for its relevance to different data types and domains. We utilize a Python-based crawling framework, which is detailed in our publicly available repository. This framework is designed to be adaptable, allowing for the inclusion of additional sources as they become relevant.

The primary data sources include:
\begin{itemize}
    \item \textbf{Financial News}: Crawled from Yahoo, providing up-to-date information on global financial trends.
\vspace{-3mm}
    \item \textbf{Political News}: BBC News is used to gather the latest political developments worldwide.
\vspace{-3mm}
    \item \textbf{Discussion Forums}: Reddit is scraped to capture current discussions across a variety of tofigures.
\vspace{-3mm}
    \item \textbf{Online Literature}: Wattpad offers a rich source of contemporary fiction and non-fiction.
\vspace{-3mm}
    \item \textbf{Encyclopedia}: Wikipedia's latest changes feed is monitored for updates across all domains.
\vspace{-3mm}
    \item \textbf{Academic Papers}: arXiv is used to access the newest research across multiple scientific fields.
\vspace{-3mm}
    \item \textbf{Code Repositories}: GitHub Trending provides insights into the latest developments in software engineering.
\vspace{-3mm}
    \item \textbf{Question and Answer Forums}: Quora, focusing on popular topics such as Technology, Mathematics, Health, and Movies.
\end{itemize}

To facilitate efficient and effective data collection, our system's environment is configured with a specific Python package index URL and utilizes Playwright for web navigation and content extraction. This setup addresses potential challenges, such as page refresh requirements on Yahoo and hostname verification issues encountered with certain websites.

Our system employs a combination of tools and libraries, including Requests, Playwright, and various PDF processing utilities, to collect the latest textual data from the internet. This approach enables us to dynamically update our benchmark datasets with fresh information, ensuring that our evaluation reflects current language use and information trends.

\textbf{Data Pre-processing}
For some raw content clean the text by removing these tags to ensure that only the content text is analyzed.

Filtering out short texts: Extremely short texts may not provide enough contextual information for analysis or model training. Setting a minimum text length threshold of 100 chars to filter out such instances.

\textbf{External Dataset}
For the analysis of text likelihood and bias, we employed not only our own collection of historical arXiv data but also sampled entries from two additional datasets. We incorporated data from the BBC News and Wikitext datasets, which are publicly available through the Hugging Face datasets repository. Specifically, the datasets used were \textit{bbc\_news\_alltime} and \textit{wikitext\_alltime}, accessible via the following URLs:

\begin{itemize}
    \item BBC News: \url{https://huggingface.co/datasets/RealTimeData/bbc_news_alltime}
    \item Wikitext: \url{https://huggingface.co/datasets/RealTimeData/wikitext_alltime}
\end{itemize}

Sampling from these datasets was carried out every three months, with each instance involving the collection of 50 entries per dataset. This structured sampling approach was designed to ensure a consistent and representative analysis of content over time, enabling an effective assessment of textual bias and likelihood trends.

\begin{table}[!h]
\centering
\caption{Most common tags of the Gjo questions}
\label{tab:gjo_tag_count}
\begin{tabular}{@{}lc@{}}
\toprule
Topic & Count \\ \midrule
Business & 648 \\
Non-US Politics & 453 \\
Security and Conflict & 446 \\
Society & 425 \\
US Politics & 363 \\
Technology & 359 \\
Finance & 299 \\
Elections and Referenda & 288 \\
Economic Indicators & 273 \\
Foreign Policy & 271 \\
US Policy & 247 \\
Health & 245 \\
Leader Entry/Exit & 219 \\
Economic Policy & 160 \\
Sports & 148 \\
Entertainment & 148 \\
Environment & 80 \\
Open & 16 \\ \bottomrule
\end{tabular}

\end{table}

\section{ Data Overview of One Crawl}
\label{app:data_overview}

Tab.~\ref{table:text_sources_details} presented below offers a comprehensive overview of the diverse sources and categories of texts that were analyzed in one data crawl. Classified into distinct groups such as Academic STEM, Academic Non-STEM, Internet QA, and Internet articles, the table delineates the number of texts extracted from each source along with their average length. This includes detailed counts from different academic archives like arXiv, various categories from Internet question and answer platforms like Reddit and Quora, and articles from well-known online platforms including Wiki and BBC.

\section{Detailed results of BPC generalization on Wiki and BBC dataset}
\label{appendix:ppl_generalization_wiki&BBC}
See Table \ref{table:ppl_generalization_bbc}.
\begin{table*}[!ht]
\caption{
Percentage Change in BPC Following Release, measured by BBC news data from 2020 to 2024.3. Dash ("-") symbols in the table indicate missing data points due to the later release dates of some models, which means there is no available data for those specific time intervals.}
\label{table:ppl_generalization_bbc}
\vspace{1mm}
\centering
\small
\begin{tabular}{lp{15.5mm}p{15.5mm}p{15.5mm}p{15.5mm}p{15.5mm}p{15.5mm}}
\toprule
\parbox{8mm}{\centering Model} & \parbox{10mm}{\centering Base BPC} & \parbox{14mm}{\centering BPC Change at 3 months (\%)} & \parbox{14mm}{\centering BPC Change at 6 months (\%)} & \parbox{14mm}{\centering BPC Change at 9 months (\%)} & \parbox{14mm}{\centering BPC Change at 12 months (\%)} & {\centering mean} \\
\midrule

OPT-13B & 0.450 & 2.273 & -2.087 & 1.265 & 0.839 & 0.572 \\
OPT-2.7B & 0.475 & 1.970 & -2.228 & 1.192 & 1.143 & 0.519 \\
LLaMA-7B & 0.450 & 1.270 & 0.177 & -7.776 & 3.266 & -0.766 \\
Pythia-12B & 0.480 & -0.233 & -3.888 & -2.612 & -1.012 & -1.936 \\
Falcon-rw-1B & 0.510 & 0.459 & -7.814 & 1.201 & -1.474 & -1.907 \\
Baichuan-13B-Chat & 0.467 & -3.645 & -1.078 & 0.413 & - & -1.436 \\
LLaMA-2-13B & 0.416 & -6.928 & 3.502 & 1.683 & - & -0.581 \\
LLaMA-2-7B & 0.434 & -7.314 & 3.092 & 1.279 & - & -0.981 \\
LLaMA-2-7B-Chat & 0.505 & -8.308 & 2.695 & 0.664 & - & -1.650 \\
Zhongjing-Base & 0.465 & -8.488 & 2.662 & -1.290 & - & -2.372 \\
InternLM-Chat-7B & 0.515 & -2.476 & -1.344 & 0.236 & - & -1.195 \\
Baichuan2-7B-Base & 0.451 & -3.059 & 0.595 & -4.649 & - & -2.371 \\
Baichuan2-7B-Chat & 0.517 & -3.571 & -0.038 & -5.315 & - & -2.975 \\
Mistral-7B-v0.1 & 0.398 & 1.110 & 5.347 & -0.661 & - & 1.932 \\
Phi-1.5 & 0.602 & -1.772 & 1.131 & -1.837 & - & -0.826 \\
Baichuan2-13B-Base & 0.439 & -2.943 & 0.882 & -4.390 & - & -2.150 \\
Baichuan2-13B-Chat & 0.495 & -3.541 & -0.995 & 0.232 & - & -1.435 \\
Colossal-LLaMA-2-7B-Base & 0.618 & 2.825 & 3.715 & - & - & 3.270 \\
Qwen-14B-Chat & 0.445 & 2.682 & 4.847 & - & - & 3.764 \\
Qwen-7B & 0.441 & 1.892 & 3.689 & - & - & 2.791 \\
Qwen-7B-Chat & 0.466 & 1.809 & 3.606 & - & - & 2.708 \\
Skywork-13B-Base & 0.429 & 7.163 & 3.261 & - & - & 5.212 \\
ChatGLM3-6B & 0.832 & 5.567 & 1.684 & - & - & 3.626 \\
Zephyr-7B-beta & 0.435 & 7.340 & 3.928 & - & - & 5.634 \\
Yi-6B & 0.427 & 6.478 & 3.144 & - & - & 4.811 \\
Yi-6B-Chat & 0.446 & 3.030 & -2.798 & - & - & 0.116 \\
Qwen-1.8B & 0.520 & 1.899 & -2.082 & - & - & -0.091 \\
Qwen-1.8B-Chat & 0.599 & 1.635 & -2.033 & - & - & -0.199 \\
RWKV-v5-Eagle-7B & 0.440 & 4.956 & 2.278 & - & - & 3.617 \\
TinyLLaMA-1.1B-Chat-v0.6 & 0.513 & 2.539 & -3.678 & - & - & -0.569 \\
Phi-2 & 0.487 & 1.882 & - & - & - & 1.882 \\
\bottomrule
\end{tabular}

\end{table*}

\newpage

\begin{table*}[!h]
\caption{Percentage Change in Language Likelihood Following Release, measured by wiki data from 2020 to 2024.3. Dash ("-") symbols in the table indicate missing data points due to the later release dates of some models, which means there is no available data for those specific time intervals.}
\label{table:ppl_generalization_wiki}
\vspace{1mm}
\centering
\small
\begin{tabular}{lcccccc}
\toprule
Model & \parbox{12mm}{\centering Base BPC} & \parbox{14mm}{\centering BPC Change at 3 months (\%) }& \parbox{14mm}{\centering BPC Change at 6 months (\%)} & \parbox{14mm}{\centering BPC Change at 9 months (\%)} & \parbox{14mm}{\centering BPC Change at 12 months (\%)} & {\centering mean} \\
\midrule

   OPT-13B & 0.075 & -0.623 & 0.105 & -0.413 & -0.818 & -0.437 \\
OPT-2.7B & 0.082 & -0.658 & 0.024 & -0.528 & -0.915 & -0.519 \\
LLaMA-7B & 0.052 & 0.214 & -0.425 & 0.379 & -0.067 & 0.025 \\
Pythia-12B & 0.071 & -0.578 & -0.933 & -0.859 & -0.903 & -0.818 \\
Falcon-rw-1B & 0.083 & -1.272 & -1.113 & -1.131 & -1.207 & -1.181 \\
Baichuan-13B-Chat & 0.058 & 0.019 & 0.124 & -0.125 & - & 0.006 \\
LLaMA-2-13B & 0.041 & 1.700 & 1.446 & 1.796 & - & 1.647 \\
LLaMA-2-7B & 0.050 & 0.774 & 0.349 & 0.517 & - & 0.547 \\
LLaMA-2-7B-Chat & 0.060 & 0.132 & -0.186 & -0.093 & - & -0.049 \\
Zhongjing-Base & 0.050 & 0.756 & 0.269 & 0.569 & - & 0.531 \\
InternLM-Chat-7B & 0.080 & -0.482 & -0.301 & -0.250 & - & -0.344 \\
Baichuan2-7B-Base & 0.059 & 0.523 & 0.157 & 0.185 & - & 0.289 \\
Baichuan2-7B-Chat & 0.068 & 0.343 & -0.035 & -0.073 & - & 0.078 \\
Mistral-7B-v0.1 & 0.053 & 0.381 & -0.028 & 0.018 & - & 0.124 \\
Phi-1.5 & 0.098 & 0.066 & -0.125 & -0.248 & - & -0.102 \\
Baichuan2-13B-Base & 0.053 & 1.019 & 0.797 & 0.900 & - & 0.905 \\
Baichuan2-13B-Chat & 0.062 & 0.299 & 0.628 & 0.512 & - & 0.480 \\
Colossal-LLaMA-2-7B-Base & 0.065 & -0.042 & -0.199 & - & - & -0.121 \\
Qwen-14B-Chat & 0.049 & 1.301 & 1.939 & - & - & 1.620 \\
Qwen-7B & 0.061 & 0.428 & 0.430 & - & - & 0.429 \\
Qwen-7B-Chat & 0.065 & 0.302 & 0.226 & - & - & 0.264 \\
Skywork-13B-Base & 0.050 & 0.075 & 0.276 & - & - & 0.175 \\
ChatGLM3-6B & 0.105 & -0.187 & -0.245 & - & - & -0.216 \\
Zephyr-7B-beta & 0.059 & -0.049 & -0.033 & - & - & -0.041 \\
Yi-6B & 0.058 & -0.071 & 0.022 & - & - & -0.024 \\
Yi-6B-Chat & 0.060 & -0.028 & 0.023 & - & - & -0.003 \\
Qwen-1.8B & 0.080 & -0.131 & -0.251 & - & - & -0.191 \\
Qwen-1.8B-Chat & 0.091 & -0.331 & -0.496 & - & - & -0.414 \\
RWKV-v5-Eagle-7B & 0.065 & 0.149 & 0.220 & - & - & 0.184 \\
TinyLLaMA-1.1B-Chat-v0.6 & 0.070 & -0.332 & -0.473 & - & - & -0.403 \\
Phi-2 & 0.072 & -0.037 & - & - & - & -0.037 \\
\bottomrule
\end{tabular}

\end{table*}

\newpage

\section{TBI of Models across the arXiv, BBC, and Wiki datasets }
\label{BPC_on_three_text_datasets}
See Table \ref{tab:complete_model_list}.
\begin{table*}[h]
\caption{Comprehensive list of models with their release dates and metrics for TBI (multiplied by 1000) across the arXiv, BBC, and Wiki datasets. Negative TBI values indicate a trend of increasing BPC, which refers to decreasing performance on MMLU.}
\label{tab:complete_model_list}
\vspace{1mm}
\centering
\setlength{\tabcolsep}{3pt} 
\renewcommand{\arraystretch}{1.2} 
\small 
\begin{tabular}{p{40mm}cccc} 
\toprule
\textbf{\centering Model Name} &
\textbf{\parbox{18mm}{\centering Released Date}} & 
\textbf{\parbox{16mm}{\centering TBI*1000 (arXiv)}} &
\textbf{\parbox{16mm}{\centering TBI*1000 (BBC)}}  &
\textbf{\parbox{16mm}{\centering TBI*1000 (Wiki)}} \\
\midrule
OPT-13B  & May 2022 & -19.4 & 1689.0 & -30.2 \\
OPT-2.7B  & May 2022 & -18.8 & 1080.7 & -46.9 \\
LLaMA-7B  & Feb 2023 & -15.3 & 713.0 & 52.0 \\
Pythia-12B  & Mar 2023 & -8.1 & 802.7 & -11.1 \\
Falcon-rw-1B  & Apr 2023 & -26.8 & 607.4 & -47.7 \\
Baichuan-13B-Chat  & Jun 2023 & -11.7 & 685.7 & 4.8 \\
Baichuan-7B-Chat  & Jun 2023 & -14.6 & 726.6 & -16.5 \\
LLaMA-2-13B  & Jul 2023 & -11.1 & 769.6 & 65.9 \\
LLaMA-2-7B  & Jul 2023 & -12.6 & 684.6 & 23.2 \\
Baichuan-13B-Chat  & Jul 2023 & -7.0 & 717.2 & 12.1 \\
Zhongjing-Base  & Jul 2023 & -15.0 & 759.7 & 56.9 \\
InternLM-Chat-7B  & Jul 2023 & -19.3 & 574.0 & -22.0 \\
Baichuan2-7B-Base  & Aug 2023 & -9.5 & 687.5 & 3.7 \\
Baichuan2-7B-Chat  & Aug 2023 & -8.4 & 781.8 & -11.6 \\
Mistral-7B-v0.1  & Sep 2023 & -15.1 & 756.8 & 10.4 \\
Phi-1.5  & Sep 2023 & -26.0 & 1212.9 & -60.7 \\
Baichuan2-13B-Base  & Sep 2023 & -9.1 & 747.9 & 15.8 \\
Baichuan2-13B-Chat  & Sep 2023 & -7.0 & 717.2 & 12.1 \\
Colossal-LLaMA-2-7B-Base  & Sep 2023 & -21.2 & 696.3 & -36.2 \\
Qwen-14B-Chat  & Sep 2023 & -7.9 & 762.2 & 51.5 \\
Qwen-7B  & Sep 2023 & -11.7 & 662.4 & 7.4 \\
Qwen-7B-Chat  & Sep 2023 & -13.2 & 705.7 & -3.6 \\
Skywork-13B-Base  & Oct 2023 & -16.3 & 549.1 & 30.8 \\
ChatGLM3-6B  & Oct 2023 & -25.5 & 412.6 & -70.4 \\
Zephyr-7B-beta  & Oct 2023 & -21.1 & 506.9 & 3.7 \\
Yi-6B  & Nov 2023 & -9.4 & 451.7 & 24.6 \\
Yi-6B-Chat  & Nov 2023 & -9.4 & 506.5 & 20.2 \\
Qwen-1.8B  & Nov 2023 & -22.7 & 374.0 & -46.2 \\
Qwen-1.8B-Chat  & Nov 2023 & -25.3 & 426.8 & -50.3 \\
RWKV-v5-Eagle-7B  & Nov 2023 & -12.8 & 511.3 & 4.2 \\
TinyLLaMA-1.1B-Chat-v0.6 & Dec 2023 & -25.0 & 300.1 & -45.3 \\
Phi-2  & Dec 2023 & -21.2 & 669.3 & -44.7 \\

\bottomrule
\end{tabular}
\end{table*}

\begin{table*}[!h]
\caption{Overview of text categories with their sources counts, and average lengths in one sample crawl.}
\label{table:text_sources_details}
\vspace{1mm}
\centering
\begin{tabular}{llcr}
\toprule
\textbf{Category} & \textbf{Source} & \textbf{Count} & \textbf{ Length} \\ \midrule

\multirow{5}{*}{Academic STEM} 
 & arXiv Mathematics & 50 & 42850 \\[2pt]
 & arXiv Computer Science & 50 & 43336 \\[2pt]
 & arXiv Physics & 50 & 42580 \\[2pt]
 & arXiv Statistics & 50 & 42619 \\[2pt]
 & arXiv Electrical Engineering and Systems Science & 50 & 43027 \\[2pt]
 \hline

\multirow{3}{*}{Academic Non-STEM} 

 & arXiv Economics & 50 & 42592 \\[2pt]
 & arXiv Quantitative Finance & 50 & 43969 \\[2pt]
 & arXiv Quantitative Biology & 50 & 42974 \\[2pt]
 \hline

Code & GitHub & 30 & 21849  \\[2pt] 
\hline

\multirow{5}{*}{Internet QA} 
 & Reddit & 583 & 643  \\[2pt]
 & Quora Health & 20 & 1150  \\[2pt]
 & Quora Movies & 69 & 1349  \\[2pt]
 & Quora Mathematics & 70 & 1365 \\[2pt]
 & Quora Technology & 80 & 1674  \\[2pt]
 \hline

\multirow{4}{*}{Internet article} 
 & Wiki & 220 & 4091  \\[2pt]
 & Wattpad & 38 & 6615  \\[2pt]
 & BBC & 45 & 4792 \\[2pt]
 & Yahoo & 41 & 5373  \\[2pt]

\bottomrule

\end{tabular}

\end{table*}

\section{Gjo data samples and most common tags of the Gjo questions}
\label{gjo_tag_count}
Gjo data samples see Table \ref{tab:gjo_tag_count} and Gjo questions see Table \ref{tab:questions}.

\begin{table*}[h!]
\centering
\caption{Sample of Questions and Their Respective Answers (Correct Answer in Bold): The percentages represent the proportion of forecasters' votes on a prediction website.}
\label{tab:questions}
\begin{tabular}{|p{3cm}|p{1.2cm}|p{1.2cm}|p{9cm}|}
\hline
\textbf{Question} & \textbf{Start Time} & \textbf{End Time} & \textbf{Possible Answers} \\ \hline
What will be the 12-month percentage change in the US Consumer Price Index (CPI) for September 2024? & May 31, 2024 05:00PM UTC & Oct 01, 2024 07:01AM UTC & \begin{tabular}[c]{@{}l@{}}Up by less than 1.200\% or down: 0\%\\ Up by at least 1.200\%, but less than 1.800\%: 0\%\\ Up by at least 1.800\%, but less than 2.400\%: 53\%\\ \textbf{Up by at least 2.400\%, but less than 3.000\%: 44\%}\\ Up by at least 3.000\%, but less than 3.600\%: 3\%\\ Up at least 3.600\%, but less than 4.200\%: 0\%\\ Up by 4.200\% or more: 0\%\end{tabular} \\ \hline
If Israeli Defense Forces (IDF) ground forces invade the Gaza Strip before 7 November 2023, when will Israel publicly announce or acknowledge that IDF ground forces have left the Gaza Strip? & Oct 17, 2023 05:00PM UTC & Oct 10, 2024 07:01AM UTC & \begin{tabular}[c]{@{}l@{}}Before 30 November 2023: 0\%\\ Between 30 November 2023 and 13 January 2024: 0\%\\ Between 14 January 2024 and 13 March 2024: 0\%\\ Between 14 March 2024 and 11 June 2024: 0\%\\ Between 12 June 2024 and 9 October 2024: 2\%\\ \textbf{Not before 10 October 2024: 96\%}\\ IDF ground forces will not invade the Gaza Strip \\before 7 November 2023: 2\%\end{tabular} \\ \hline
What will Kamala Harris' favorability rating be as of 3 October 2024, according to FiveThirtyEight? & Sep 17, 2024 08:00AM UTC & Oct 03, 2024 07:01AM UTC & \begin{tabular}[c]{@{}l@{}}Lower than 42.0\%: 1\%\\ At least 42.0\%, but less than 44.0\%: 4\%\\ At least 44.0\%, but less than 46.0\%: 7\%\\ \textbf{At least 46.0\%, but less than 48.0\%: 70\%}\\ At least 48.0\%, but less than 50.0\%: 15\%\\ At least 50.0\%, but less than 52.0\%: 2\%\\ At least 52.0\%, but less than 54.0\%: 1\%\\ 54.0\% or higher: 0\%\end{tabular} \\ \hline
\end{tabular}

\end{table*}

\section{Analysis of Language Likelihood Values and Benchmark Correlation Across Text and Model Groups}
\label{app:corr_length}

\begin{figure*}[h]
    \centering
    \begin{subfigure}[b]{0.32\textwidth}
        \includegraphics[width=\textwidth]{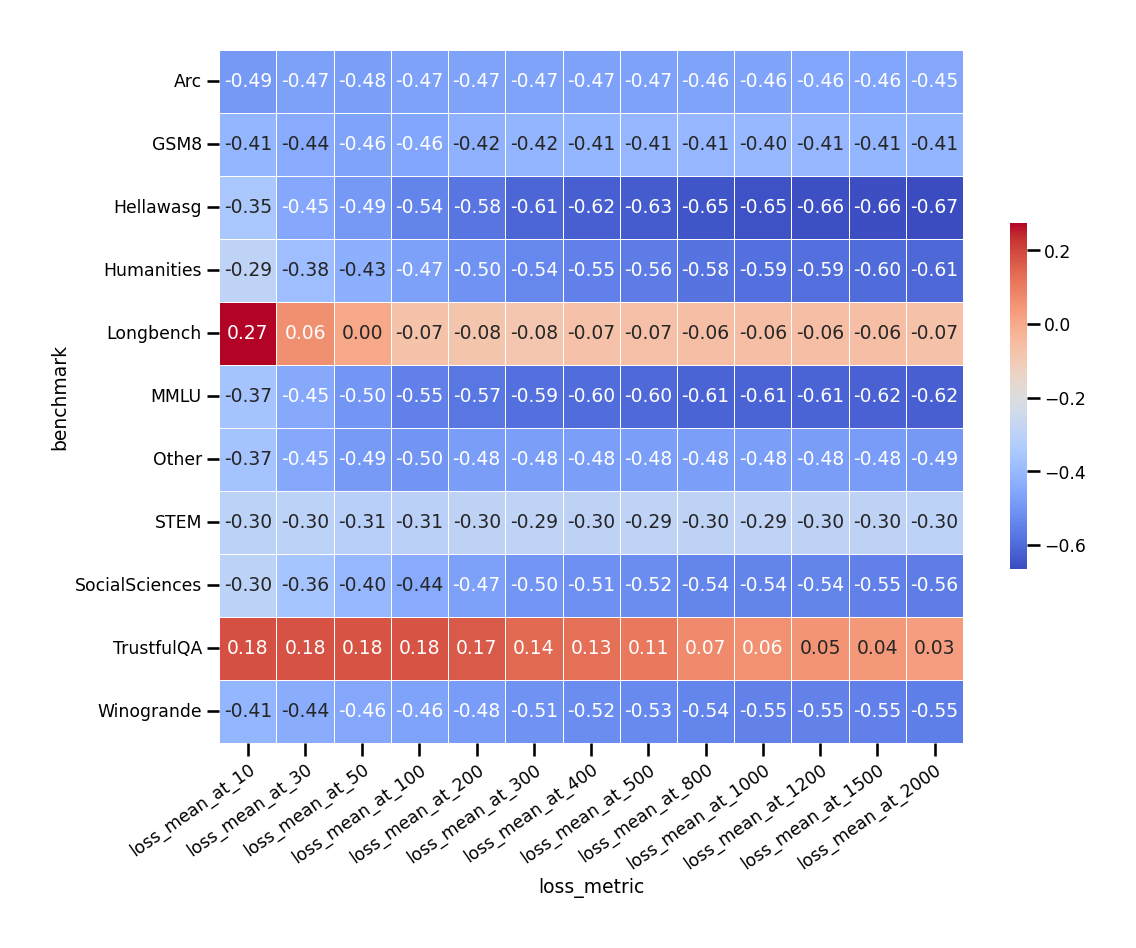}
        \caption{STEM: base models}
        \label{fig:1}
    \end{subfigure}
    \begin{subfigure}[b]{0.32\textwidth}
        \includegraphics[width=\textwidth]{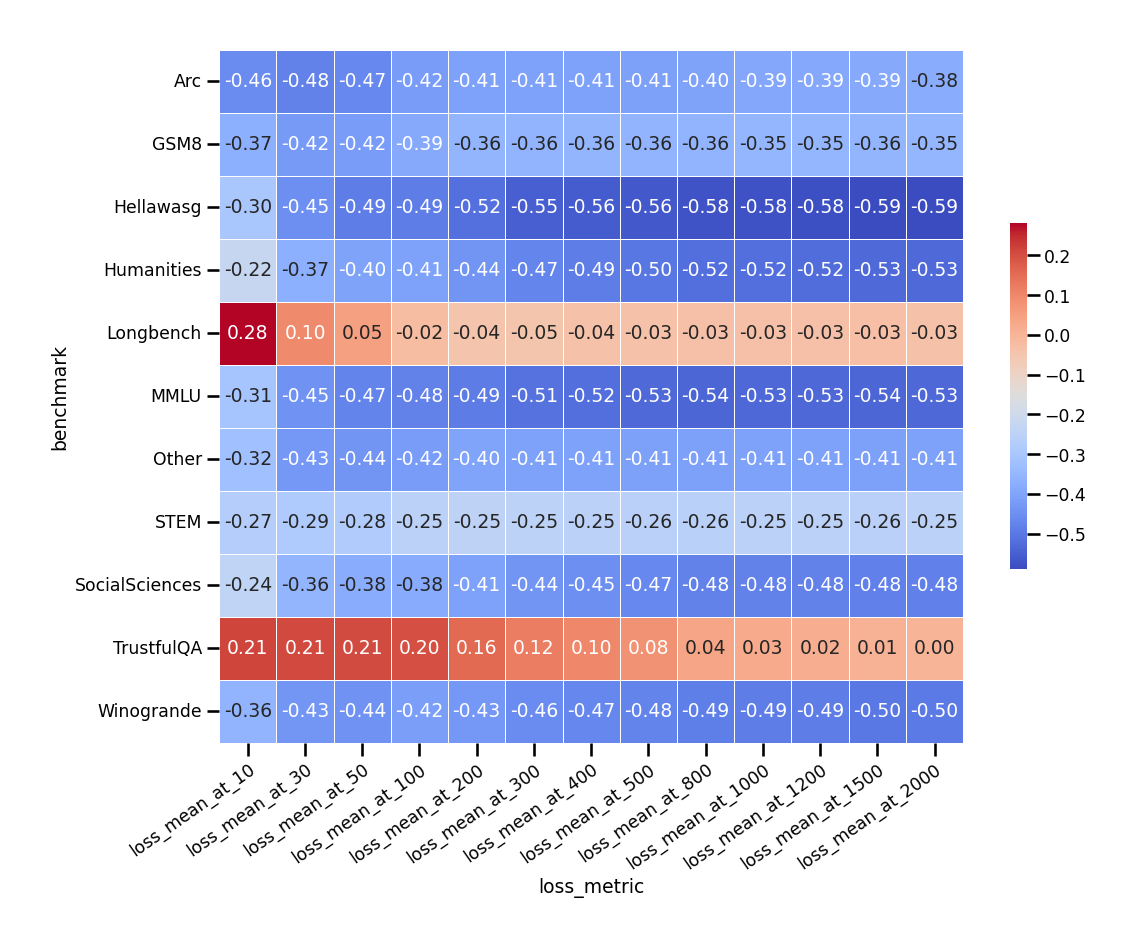}
        \caption{non-STEM: base models}
        \label{fig:2}
    \end{subfigure}
    \begin{subfigure}[b]{0.32\textwidth}
        \includegraphics[width=\textwidth]{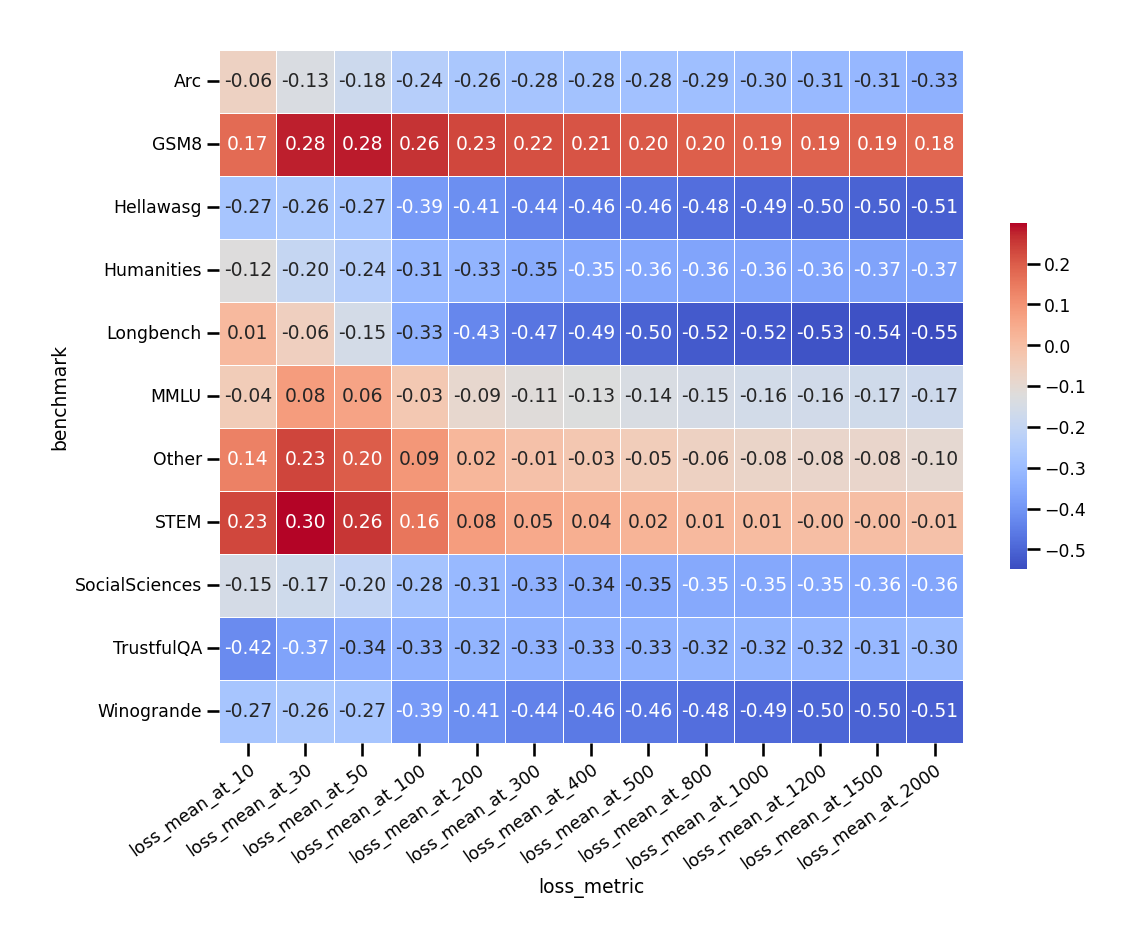}
        \caption{Non-STEM: chat models}
        \label{fig:3}
    \end{subfigure}
    
    \caption{The heatmaps display the correlations by loss for a group of models across different text chunk lengths and different benchmark scores. Each row represents a specific truncation length, while the columns represent different benchmark score categories. The values in the cells indicate the corresponding correlation coefficients.
    (a) presents the correlations by loss for all 18 base models on STEM Academia texts.
    (b) presents the correlations by loss for all 18 base models on non-STEM Academia texts.
    (c) presents the correlations by loss for all 15 chat models on non-STEM Academia texts.
    }
    \label{fig:heatmap_academic}
\end{figure*}

\begin{figure*}[h]
    \centering
    \begin{subfigure}[b]{0.32\textwidth}
        \includegraphics[width=\textwidth]{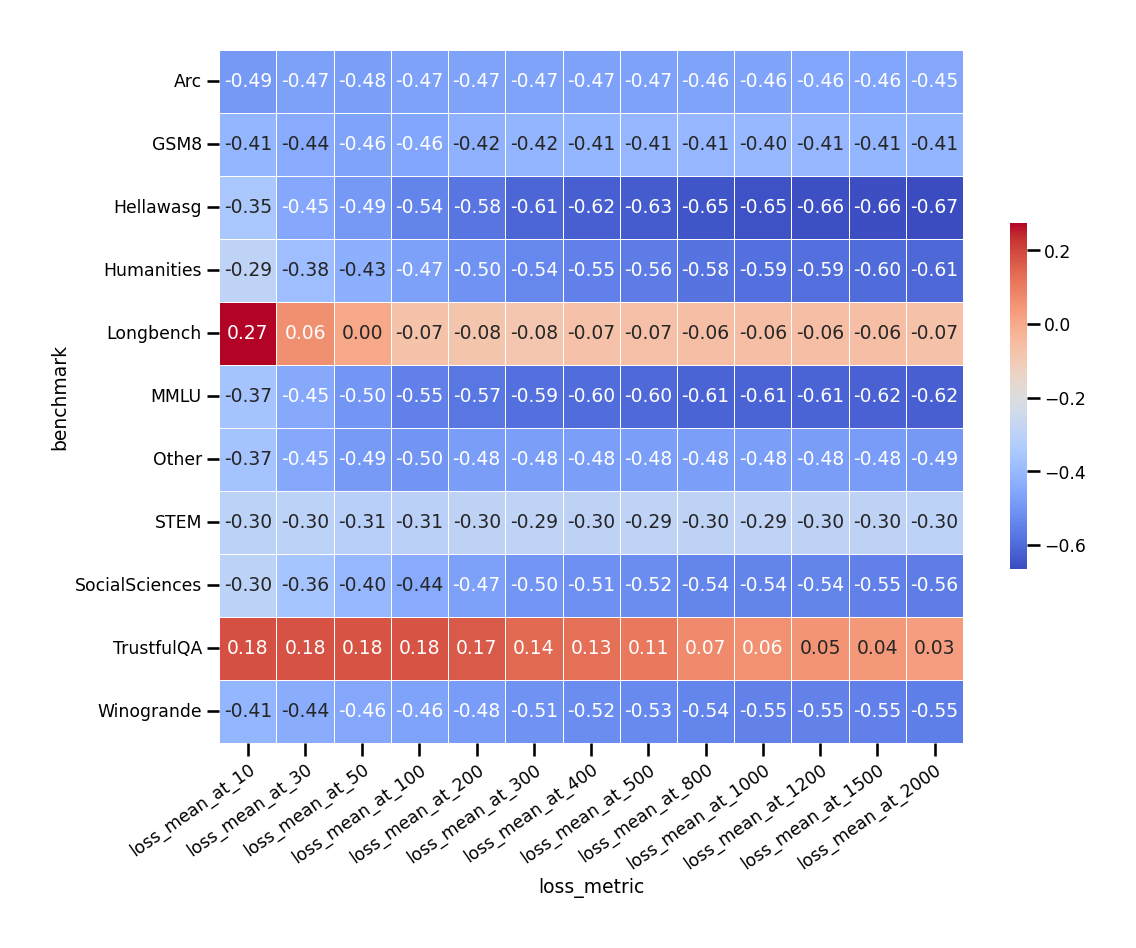}
        \caption{STEM: base models}
        \label{fig:1}
    \end{subfigure}
    \begin{subfigure}[b]{0.32\textwidth}
        \includegraphics[width=\textwidth]{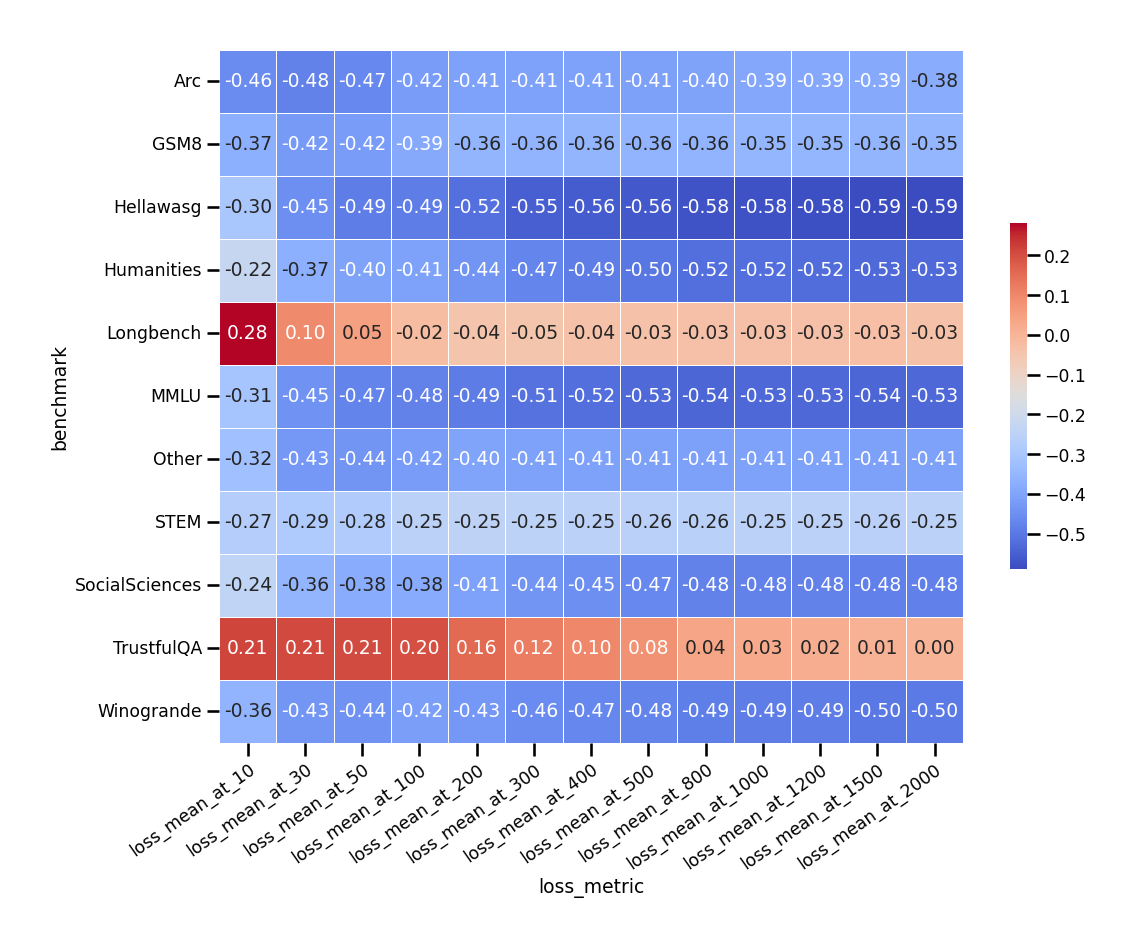}
        \caption{non-STEM: base models}
        \label{fig:2}
    \end{subfigure}
    \begin{subfigure}[b]{0.32\textwidth}
        \includegraphics[width=\textwidth]{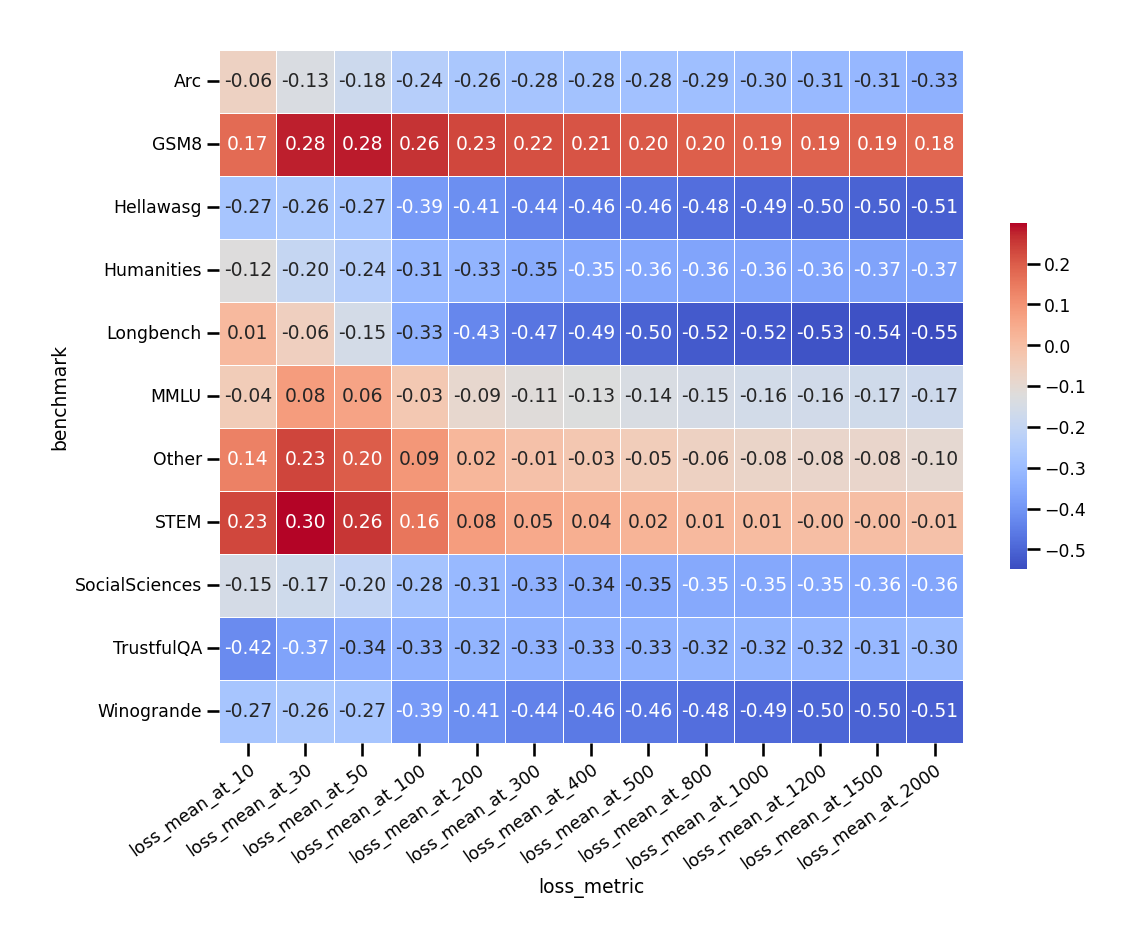}
        \caption{Non-STEM: chat models}
        \label{fig:3}
    \end{subfigure}
    
    \caption{The heatmaps display the correlations by loss for a group of models across different text chunk lengths and different benchmark scores. Each row represents a specific truncation length, while the columns represent different benchmark score categories. The values in the cells indicate the corresponding correlation coefficients.
    (a) presents the correlations by loss for all 18 base models on STEM Academia texts.
    (b) presents the correlations by loss for all 18 base models on non-STEM Academia texts.
    (c) presents the correlations by loss for all 15 chat models on non-STEM Academia 
    }
    \label{fig:heatmap_academic}
\end{figure*}

\begin{figure*}[h!]
    \centering
    \begin{subfigure}[b]{0.49\textwidth}
        \includegraphics[width=\textwidth]{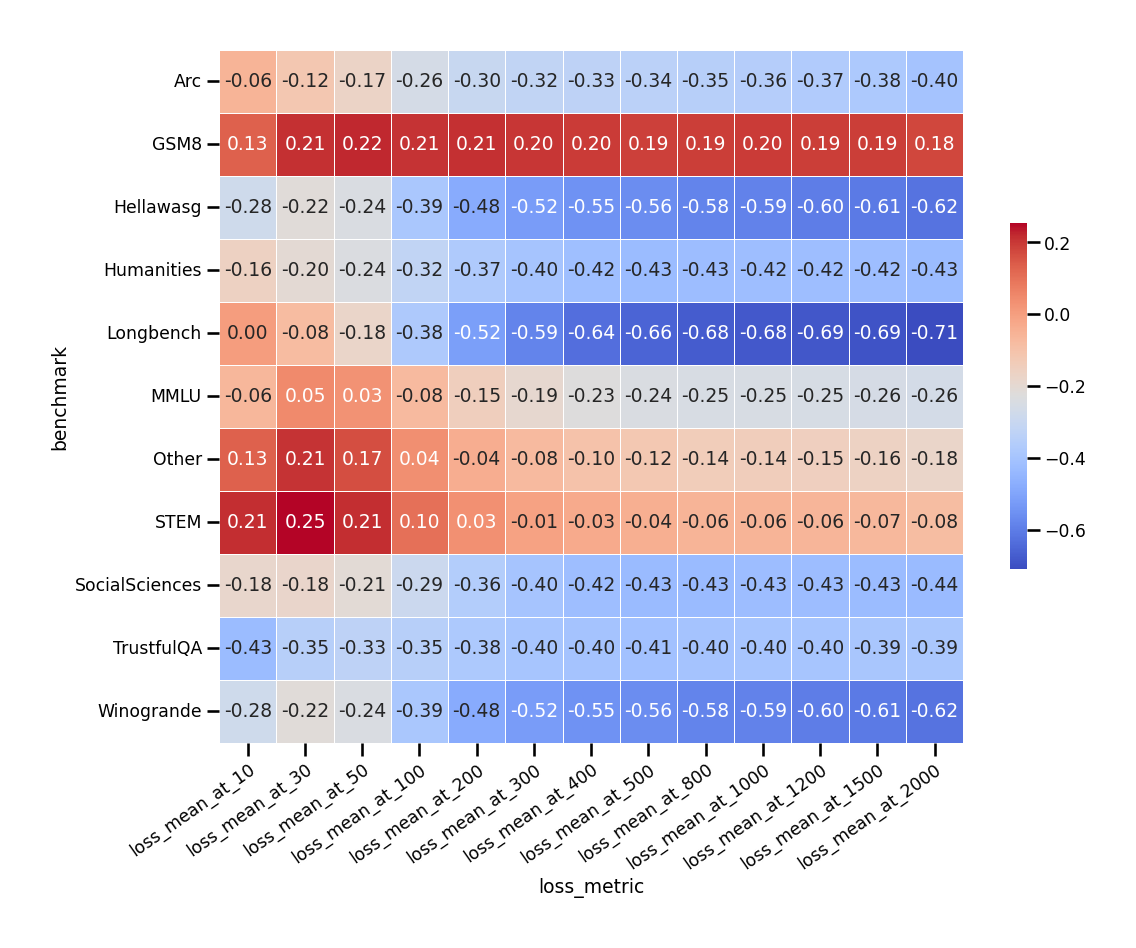}
        \caption{STEM: all chat models}
        \label{fig:1}
    \end{subfigure}
    \hfill 
    \begin{subfigure}[b]{0.49\textwidth}
        \includegraphics[width=\textwidth]{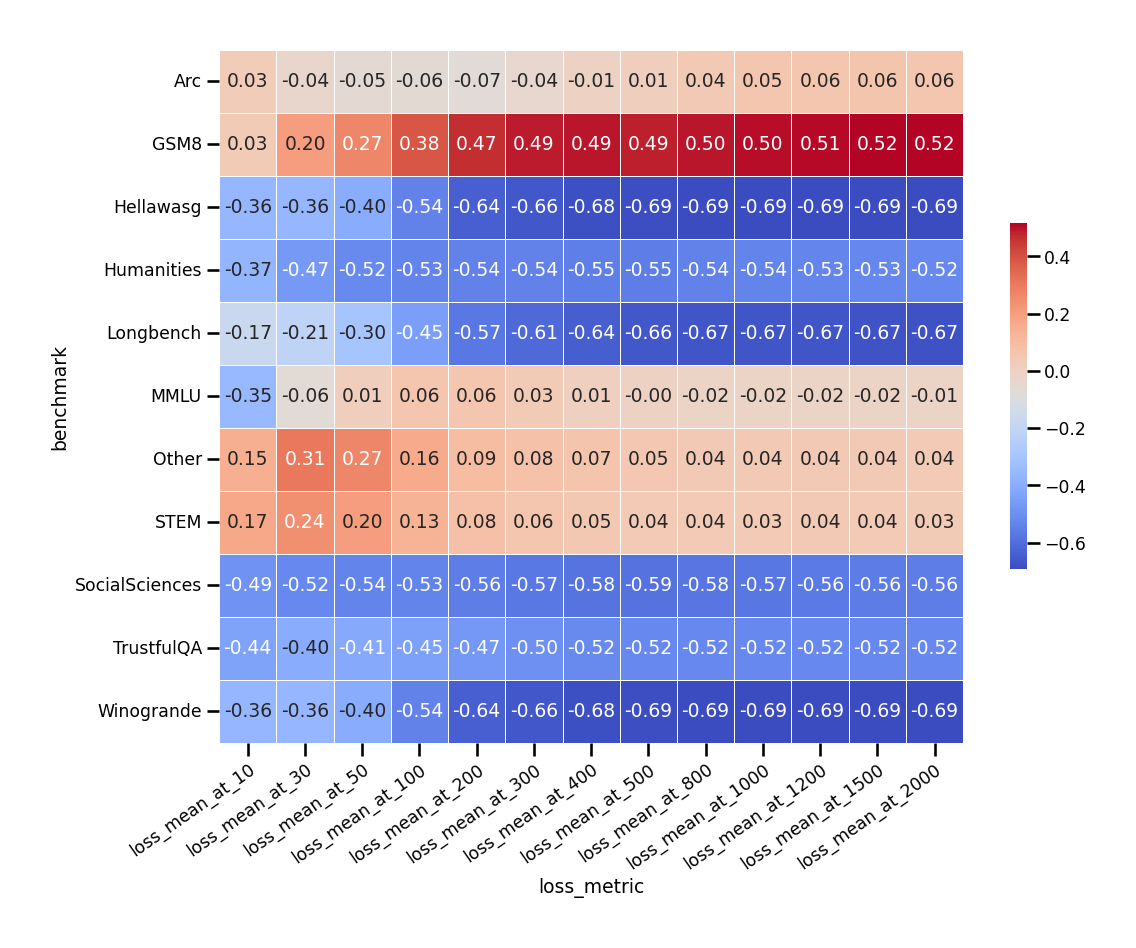}
        \caption{STEM: 7B level chat models}
        \label{fig:2}
    \end{subfigure}
    
    \caption{(a) presents the correlations by loss for all 15 chat models on STEM Academia texts.
    (b) presents the correlations by loss for all 9 7B level chat models on STEM Academia texts.
    }
    \label{fig:heatmap_model_size}
\end{figure*}

From Fig.~\ref{fig:heatmap_model_size} we can conclude that the scale of models being observed does not significantly alter outcomes, suggesting that performance correlations across benchmarks are generally consistent regardless of model size.

\section{Correlation with Existing Benchmarks}
\label{subsection:BPC Correlation}

\paragraph{The Selected Existing Benchmarks}
\label{para:selected_bench}

We explored the correlation between BPC and existing benchmarks, particularly focusing on how specific content types, such as GSM, influence model performance. The selected benchmarks encompass common sense reasoning, depth and breadth of knowledge across diverse subjects, truthfulness, natural language understanding, and mathematical reasoning, ensuring a well-rounded assessment of the model's performance.

\begin{table*}[h]
\caption{Overview of AI Benchmarks and Their Evaluation Focus}
\label{tab:ai_benchmarks}
\vspace{1mm}
\centering
\renewcommand{\arraystretch}{1.22}
\begin{tabular}{lp{11cm}}
\toprule
\textbf{Benchmark} & \textbf{Description} \\
\midrule
HellaSwag & Tests an AI's common sense reasoning by requiring it to complete sentences or narratives that reflect everyday logic, highlighting the model's ability to predict logical continuations. \citet{zellers2019hellaswag} \\ \hline
MMLU & Evaluates an AI's breadth and depth of understanding across a diverse range of subjects, from science to literature, showcasing its versatility and comprehensive grasp of human language. \citet{hendrycks2020measuring} \\  \hline
TruthfulQA & Focuses on the model's capacity to provide honest and accurate responses, particularly to questions where the truth may not be intuitive, testing its commitment to truthfulness and its proficiency in avoiding misinformation. \citet{lin2022truthfulqa} \\  \hline
Winogrande & Assesses common sense reasoning and natural language understanding through sentence completion tasks, measuring an AI's ability to apply common sense knowledge in language processing. \citet{sakaguchi2019winogrande} \\  \hline
GSM8K & Challenges AI models with elementary-level math problems, testing their mathematical reasoning capabilities and their understanding and application of basic math concepts. \citet{gsm8k_cobbe2021training} \\
\bottomrule
\end{tabular}

\end{table*}

\paragraph{Base vs. Chat Models}
Tab.~\ref{table:correlationbasechat} shows the comparative analysis of base and tuned models. Our findings suggest that:
1) Base models exhibit a stronger correlation with established benchmarks, implying that these models, in their fundamental form, possess a robust capacity for modeling language. This observation advocates for conducting initial evaluations using base models to capture their intrinsic language modeling capabilities. 2) The exploration of tuned models reveals potential deeper patterns in model tuning, suggesting that specific tuning strategies may unlock further enhancements in model performance across different tasks and domains.

\begin{table*}[!h]
\caption{Correlation with Different Benchmarks, Base vs. Chat Models}
\label{table:correlationbasechat}
\vspace{1mm}
\centering
\begin{tabular}{lccr}
\toprule
Benchmarks & Loss of Base Model  & Loss of Chat Model & Delta \\
\midrule
MMLU & -0.61 & -0.22 & 0.39 \\
MMLU:Humanities & -0.59 & -0.42 & 0.17 \\
MMLU:SocialSciences & -0.54 & 0.42 & 0.96 \\
MMLU:STEM & -0.29 & 0.02 & 0.31 \\
MMLU:Other & -0.47 & -0.1 & 0.37 \\
Arc & -0.46 & 0.34 & 0.8 \\
Hellawasg & -0.66 & -0.57 & 0.09 \\
Wnogrande & -0.56 & 0.57 & 1.13 \\
Longbench & -0.04 & -0.65 & -0.61 \\
GSM8K & -0.4 & 0.22 & 0.62 \\
TrustfulQA & 0.05 & 0.39 & 0.34 \\
\bottomrule
\end{tabular}

\end{table*}

\paragraph{Correlation alone Text Length}

Tab.~\ref{table:correlations_horizontal} is about a long text understanding benchmark Longbench. 

For chat models in this task, the correlation between BPC and scores is logical—higher probabilities of generating the text correlate with higher scores, and this correlation strengthens with longer text lengths. This suggests a consistent alignment between the model’s language likelihood in long text and its scoring, reinforcing the chat model's robustness in handling extensive text inputs.

However, an inverse trend is observed with base models, where a familiar text paradoxically results in lower scores on related comprehension tasks. This indicates that fine-tuning may alter the base models' understanding of the texts, perhaps shifting their processing in ways that do not favor traditional comprehension metrics.

This indicates scenarios where a model scores high on language likelihood due to its text-generation skills but may not perform equally well on tasks that require deep semantic understanding or critical thinking. Thus, relying solely on language likelihood as an indicator of overall performance might overlook crucial aspects of cognitive and interpretative abilities that are essential for more complex applications.


\begin{table*}[!h]
\caption{Correlation with Longbench values across different lengths}
\label{table:correlations_horizontal}
\vspace{1mm}
\centering
\resizebox{\textwidth}{!}{%
\setlength\tabcolsep{3pt}
\begin{tabular}{lccccccccccccc}
\toprule
Length & 10 & 30 & 50 & 100 & 200 & 300 & 400 & 500 & 800 & 1000 & 1200 & 1500 & 2000 \\
\midrule
Base Models & 0.35 & 0.13 & 0.06 & -0.03 & -0.05 & -0.06 & -0.05 & -0.05 & -0.04 & -0.04 & -0.04 & -0.04 & -0.04 \\
Chat Models & 0.01 & -0.07 & -0.16 & -0.39 & -0.53 & -0.58 & -0.61 & -0.63 & -0.65 & -0.65 & -0.67 & -0.67 & -0.68 \\
\bottomrule
\end{tabular}%
}
\end{table*}

\section{Details of Model Accuracy Peaks}


Tab.~\ref{table:acc_peak} captures the peak performance periods of various models on Freshbench. It lists specific models along with their respective dates when they achieved the best performance.

\begin{table*}[!h]
\caption{Detailed Record of Model Accuracy Peaks
}
\label{table:acc_peak}
\vspace{1mm}
\centering
\renewcommand{\arraystretch}{1.5}
\begin{tabular}{lp{11.5cm}}
\toprule
Date & Models \\ \hline
2015-01 & Skywork-13B-Base, Phi-1.5 \\ \hline
2015-07 & Skywork-13B-Base, Phi-1.5 \\ \hline
2016-01 & Baichuan2-7B-Chat, Colossal-LLaMA-2-7B-Base, LLaMA-2-7B, Qwen-14B-Chat, Qwen-1.8B, Qwen-1.8B-Chat, Yi-6B, Yi-6B-Chat, Baichuan-13B-chat, Baichuan-7B-chat, ChatGLM3-6B, falcon-1B, InternLM-chat-7B, OPT-13B, Phi-2 \\ \hline
2016-07 & RWKV5-Eagle-7B \\ \hline
2017-01 & GPT-3.5-turbo-0613 \\ \hline
2017-07 & Baichuan2-7B-Base \\ \hline
2018-01 & Baichuan2-13B-Base \\ \hline
2018-07 & Baichuan2-13B-Chat, LLaMA-2-13B-hf, Qwen-7B, GPT4-0613, GPT4-1106, LLaMA2-7B-Chat, LLaMA-7B, mistral-7B-v0.1, OPT-2.7B, Pythia-12B, Zephyr-7B-beta \\ \hline
2019-07 & TinyLLaMA-1.1B-Chat \\ \hline
2020-01 &  \\ \hline
2020-07 &  \\ \hline
2021-01 &  \\ \hline
2021-07 &  \\ \hline
2022-01 &  \\ \hline
2022-07 &  \\ \hline
2023-01 &  \\ \hline
2023-07 &  \\ \hline
2024-01 &  \\ \hline
2024-07 &  \\ \hline
\end{tabular}

\end{table*}

\section{Analysis of Best-Achieved Accuracy and Cutoff Time}

\begin{table*}[!h]
\caption{Analysis about the time of the best-achieved accuracy and its cutoff time.  }
\label{table:model_time_segments}
\vspace{1mm}
\centering
\footnotesize 
\setlength{\tabcolsep}{3pt} 
\renewcommand{\arraystretch}{1.2} 
\begin{tabular}{llcccllllll}
\toprule
Peak Period & Model & Peak Time of Accuracy & Cutoff Time & Interpolation \\
\midrule
Post-2019 & TinyLLaMA1.1B   & 2019-07 & 2023-11 &  52 \\
\midrule
 2017 to 2018 & GPT4-1106 & 2018-07 & 2023-06 &  59 \\
 & LLaMA2-13B & 2018-07 &  2023-02&  55 \\
 & Baichuan2-13B-Base & 2018-01 &  2023-06& 65 \\

\midrule
2015 to 2016 & Baichuan2-7B-Chat & 2016-01 & 2023-08 & 91  \\
 & Colossal-LLaMA-2-7B-Base & 2016-01 & 2023-09 & 92 \\
 & LLaMA2-7B & 2016-01 & 2023-02 & 85 \\
\bottomrule
\end{tabular}
\end{table*}

Tab.~\ref{table:model_time_segments} shows an analysis covering the period from 2015 to 2024, with the year 2019 serving as a pivotal midpoint, a distinct pattern emerges from the performance data of various language models: all models reached their peak accuracy before 2020. This observation points to a notable trend toward "\textcolor{navy}{Nostalgia}", where models demonstrate optimal performance on data from the earlier part of the timeline analyzed. 
This tendency highlights the models' alignment with the informational characteristics prevalent prior to the latter half of the observed period, suggesting their training data may disproportionately reflect earlier times.

The peak accuracy periods analyzed are based on the average accuracy of all questions within three-month intervals, to provide a stable and robust assessment of each model's performance.

\end{document}